\documentclass{article}
\usepackage{PRIMEarxiv}

\usepackage{algorithm}
\usepackage{algpseudocode}

\usepackage[utf8]{inputenc} 
\usepackage[T1]{fontenc}    
\usepackage{hyperref}       
\usepackage{url}            
\usepackage{booktabs}       
\usepackage{amsfonts}       
\usepackage{nicefrac}       
\usepackage{microtype}      
\usepackage{lipsum}
\usepackage{graphicx}

\usepackage{hyperref}
\usepackage{amsmath}
\usepackage{amssymb}
\usepackage{braket}
\usepackage{placeins}
\usepackage{makecell}
\usepackage{multirow}
\usepackage{caption}

\usepackage{colortbl}
\usepackage[dvipsnames,table,xcdraw]{xcolor}

\newcommand{\blue}[1]{\textcolor{CornflowerBlue}{#1}}
\newcommand{\green}[1]{\textcolor{ForestGreen}{#1}}
\newcommand{\bluecell}{\cellcolor{CornflowerBlue}\textcolor{white}}
\newcommand{\greencell}{\cellcolor{ForestGreen}\textcolor{white}}
 \newcommand{\bspline}[1]{\textcolor[HTML]{377eb8}{#1}}
\newcommand{\tanhc}[1]{\textcolor[HTML]{984ea3}{#1}}
\newcommand{\grbf}[1]{\textcolor[HTML]{a65628}{#1}}
\newcommand{\fourier}[1]{\textcolor[HTML]{4daf4a}{#1}}

\newcommand{\ptanhc}[1]{\textcolor[HTML]{ff7f00}{#1}}
\newcommand{\rev}[1]{\textcolor{black}{#1}}

\usepackage{pifont} 
\title{Learnable Activation Functions in Physics-Informed Neural Networks for Solving Partial Differential Equations
}

\author{
  Afrah Farea, Mustafa Serdar Celebi \\[0.5em]
  Computational Science and Engineering Department, Informatics Institute \\
  Istanbul Technical University, Istanbul 34469, Turkiye \\
  \texttt{farea16@itu.edu.tr, mscelebi@itu.edu.tr} \\  
}

\begin{document}
\maketitle

\begin{abstract}
Physics-Informed Neural Networks (PINNs) have emerged as a promising approach for solving Partial Differential Equations (PDEs). However, they face challenges related to spectral bias (the tendency to learn low-frequency components while struggling with high-frequency features) and unstable convergence dynamics (mainly stemming from the multi-objective nature of the PINN loss function).
These limitations impact their accuracy for problems involving rapid oscillations, sharp gradients, and complex boundary behaviors.
We systematically investigate learnable activation functions as a solution to these challenges, comparing Multilayer Perceptrons (MLPs) using fixed and learnable activation functions against Kolmogorov-Arnold Networks (KANs) that employ learnable basis functions.
Our evaluation spans diverse PDE types, including linear and non-linear wave problems, mixed-physics systems, and fluid dynamics.
\rev{
Using empirical Neural Tangent Kernel (NTK) analysis and Hessian eigenvalue decomposition, we assess spectral bias and convergence stability of the models.
Our results reveal a trade-off between expressivity and training convergence stability. While learnable activation functions work well in simpler architectures, they encounter scalability issues in complex networks due to the higher functional dimensionality. 
Counterintuitively, we find that low spectral bias alone does not guarantee better accuracy, as functions with broader NTK eigenvalue spectra may exhibit convergence instability. We demonstrate that activation function selection remains inherently problem-specific, with different bases showing distinct advantages for particular PDE characteristics.
}
We believe these insights will help in the design of more robust neural PDE solvers.
\end{abstract}
 
\section{Introduction}
\label{sec:Introduction}

\rev{
Solving partial differential equations (PDEs) underpins numerous scientific and engineering applications, from fluid dynamics to electromagnetic wave propagation. While traditional numerical methods like finite element analysis provide accurate solutions, they face computational bottlenecks in high-dimensional spaces and require extensive mesh 
generation. 
}
Physics-Informed Neural Networks (PINNs)~\cite{raissi2019physics} offers a mesh-free alternative that directly incorporates physical laws into the learning process. 
By embedding governing equations and boundary conditions as part of the loss function, PINNs offer a physics- and data-efficient approach for both forward and inverse PDEs, enabling the discovery of unknown physical parameters even with limited datasets~\cite{cuomo2022scientific, karniadakis2021physics, jin2021nsfnets, lagaris1998artificial, mendez2023, arzani2021uncovering, MOSTAJERAN2022108236, ZHANG2024109130}.

Achieving accurate solutions and maintaining convergence stability are ongoing challenges in PINNs. A significant issue is spectral bias~\cite{rahaman2019spectral, xu2020frequency, cai2019multi}, which leads to slow convergence
especially in wave propagation (e.g., Helmholtz, Wave, Klein-Gordon), multi-physics flows (e.g., Convection-diffusion), and in fluid dynamics (e.g., Cavity flow).
Researchers have explored alternative architectures and different activation functions, such as those mentioned in ~\cite{li2020fourier,moseley2023finite,10227556,liu2024kan,rathore2024challenges,waheed2022kronecker}, among others.
More specifically, the Kolmogorov Arnold Network (KAN)~\cite{liu2024kan} offers flexibility through different basis functions (e.g., splines or polynomials) and trainable scaling factors.
Recent studies suggest that KANs exhibit reduced spectral bias, improving their ability to capture high-frequency components~\cite{Wang2024spectral, Koenig2024:KAN-ODE:CMAME}.
Despite these advances, the spectral bias problem remains a challenge for MLPs and KANs~\cite{choraria2022spectral,wu2022extrapolation,hong2022activation,kammonen2024comparing}.

Handling PDEs with different frequencies presents a challenge for traditional numerical methods as well. A solution to this issue is the Multigrid Method (MG)~\cite{briggs2000multigrid, trottenberg2000multigrid}, which uses hierarchical grids to manage various frequency components. This method employs iterative solvers to initially smooth high-frequency components on finer grids while correcting low-frequency errors on coarser grids. In the context of neural network architectures, such as KANs, the incorporation of learnable basis functions introduces adaptive grid sizes. 
\rev{This leads to an important question:
Can learnable activation functions mitigate spectral bias and enhance convergence stability in PINNs when approximating multi-scale PDEs? Does low spectral bias indeed guarantee improved accuracy and stability, or do other factors dominate the optimization landscape? Understanding these relationships is crucial for developing robust neural PDE solvers that can effectively handle the multi-scale nature of complex physical systems.
}


In this paper, our main contributions are:

\begin{enumerate}
    \item
    \textbf{Exploration of learnable activations:} We extensively compare KANs and MLPs with both fixed and learnable activation functions (in this paper, we use the terms ``learnable'' and ``trainable'' interchangeably), revealing their respective strengths in training convergence and testing error. 

    \item
    \textbf{Study of convergence and spectral bias behavior:} Using the empirical Neural Tangent Kernel (NTK) and maximum eigenvalue of the loss Hessian, we analyze each model's capacity to approximate high-frequency patterns and convergence stability. Additionally, we investigate whether low spectral bias guarantees good testing outcomes.

    \item
    \textbf{Application to diverse PDEs:} We evaluate performance across PDEs with varying spectral properties: oscillatory problems (e.g., Helmholtz~\cite{mcclenny2023self, liuBinaryStructuredPhysicsinformed2024, dolean2024multilevel, wang2024adaptive} and Wave~\cite{bai2023physics,roy2024exact}), nonlinear wave (e.g., Klein-Gordon~\cite{jagtap2020adaptive,wang2024practical}), mixed-physics systems (e.g., Convection-diffusion~\cite{rezaei2022mixed,badia2024finite}), and complex fluid dynamics (e.g., Cavity~\cite{song2024loss,huang2024efficient,takamoto2022pdebench,yang2024data}) problems, 
    using various activation and basis functions (e.g., Tanh, parameterized-Tanh, B-splines, GRBFs, Fourier, Chebyshev, and Jacobi polynomials)
\end{enumerate}

The source code and pre-trained models used in this study are publicly available on GitHub at~\url{https://github.com/afrah/pinn_learnable_activation.git} and Zenodo~\cite{afrah2025-zenedo}, facilitating reproducibility and further exploration.

\section{Related Work}
\label{sec:related_works}
Adaptive activation functions \rev{(e.g., learnable or trainable activation function)} play a crucial role in neural networks by enhancing their expressivity and reducing the need for manual hyperparameter tuning, including the challenging domain of PINNs. However, this adaptability comes at the cost of increased training complexity, optimization difficulty, and computational overhead.

Notably,  Apicella et al.~\cite{apicella2021survey} classified learnable activation functions into two categories. The first category is parameterized standard activation functions, in which specific parameters are learned during training. For example, the original Swish activation proposed in~\cite{ramachandran2017searching} has the form $x.\sigma(\beta x)$ where, $\beta$ is a constant or learnable parameter. The second category consists of functions based on ensemble methods, where multiple candidate activation functions or models are combined to create more flexible and adaptable activation function $f(x)$ such that $f(x) = \sum_{i=1}^h \alpha_i g_i(x)$ where, $g_i(x)$ is a linear mapping, and $\alpha_i$ are learnable weights. 
Sütfeld et al.~\cite{sutfeld2020adaptive} proposed Adaptive Blending Units (ABUs) that combine a set of basic activation functions through learnable blending weights, allowing the neural network to learn an optimal activation shape for each layer during training. 
Wang et al.~\cite{wang2023learning} followed a similar approach using $g_i$ as a standard activation function and experimentally showed that this method outperforms the fixed activation function. However, the success of this method depends on the diversity and suitability of the candidate activation functions, and the complexity increases with an increasing number of candidate functions. 

Jagtap et al.~\cite{jagtap2020adaptive} demonstrated that using adaptive activation functions significantly enhances the training process. 
Liu et al.~\cite{liu2024kan} demonstrated that KAN models with B-spline basis functions can accurately solve the Poisson equation, outperforming MLPs in accuracy and parameter efficiency. 
Zang et al.~\cite{zeng2024kan} showed that KAN performs better on regular functions, but MLP generally outperforms KAN on functions with irregularities like discontinuities and singularities. 
Hong et al.~\cite{hong2022activation} developed a piecewise linear spline activation function, known as the Hat function (inspired by finite element methods), to balance the learning behavior across frequency spectra.
Recent works, such as~\cite{Wang2024spectral, Shukla2024, rigas2024adaptive, wang2024kolmogorov}, showed that KANs can improve convergence and reduce spectral bias compared to MLPs, thereby enhancing their ability to represent complex functions. 

\rev{
Despite these advances, significant limitations persist in current research. Current studies are primarily focused on straightforward PDEs or small-scale issues. Additionally, there is a lack of systematic comparisons among various types of PDEs with different spectral characteristics in the literature.}

\section{Problem Formulation and Neural Network Methods}

\rev{
Our methodology includes: 
(a) Study of PINNs with architectures such as MLPs and KANs. This section will detail these architectures.
(b) Comprehensive evaluation of these architectures with various activation functions across PDEs (in Section~\ref{sec:use-cases}) that have varying spectral characteristics. The results of this testing will be presented in Section~\ref{sec:computational-experiment} and ~\ref{sec:analysis}.
(c) Lastly, we provide a theoretical analysis of the results using NTK and Hessian eigenvalue decomposition to understand convergence behavior in Section~\ref{sec:analysis}.
}

\rev{\subsection{Differential Equation}}

\begin{figure*}[t]
    \centering
    \includegraphics[width=1.0\textwidth, trim={0.0cm 0.0cm 0.0cm 0.0cm}, clip]{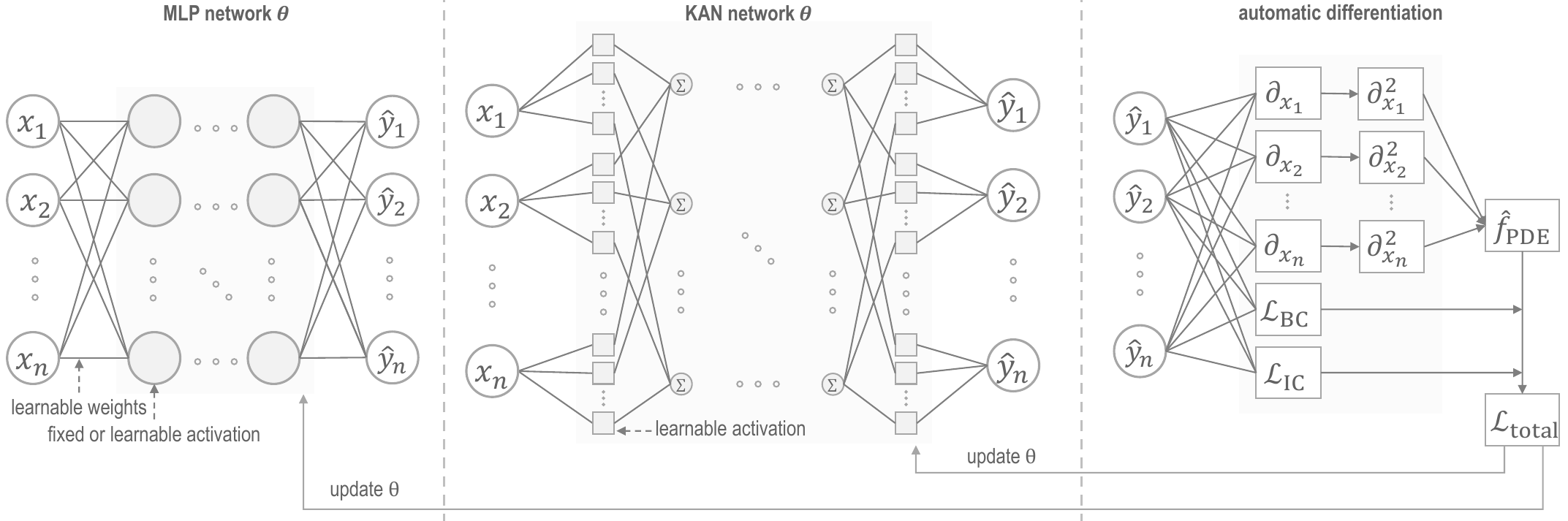}    
    \caption{This diagram presents a high-level comparison between the MLP and KAN architectures within the PINN framework. The MLP network uses learnable weights with nonlinear activation functions during training. In contrast, KAN incorporates learnable activation functions. The MLP and KAN networks pass the outputs to the Physics module, which uses automatic differentiation (AD) to calculate physics loss.
    }
    
    \label{fig:mlp-vs-kan}
\end{figure*}

The general differential equation  has the form:

\begin{align}
    \label{eq:general_pde_eq}
    \mathcal{D}[u(\mathbf{x});\alpha] &= f(\mathbf{x}) , \quad \mathbf{x} \in \Omega, 
    \\
    \mathcal{B_k}[u(\mathbf{x})] &= g_k(\mathbf{x}) , \quad \mathbf{x} \in \Gamma_k \subset \partial \Omega ,\quad k =1,2,\hdots ,n_b \notag
\end{align}

\noindent where, $\mathcal{D}$ is the differential operator, $\mathcal{B_k}$ is a set of boundary operators,  $ f(\mathbf{x})$ is the forcing function, $ g_k(\mathbf{x})$ is a set of boundary functions, $\alpha$ is a set of parameters used in the differential equation and $u(\mathbf{x})$ is the solution of the differential equation given input $\mathbf{x}$  in the solution domain $\Omega$ with boundary $\partial \Omega$. 
This work considers five PDEs as use cases discussed in Section~\ref{sec:use-cases}. 

\subsection{Multilayer Perceptron Representation}

\rev{
We start with standard MLPs as our baseline.
}
Fig.~\ref{fig:mlp-vs-kan}(left) shows the structure of an MLP model with learnable weights and activation functions (fixed or learnable).  
The effectiveness of MLPs is theoretically grounded in the universal
approximation theorem~\cite{cybenko1989approximation}, which asserts that a feedforward network with a single hidden layer containing $k > N(\epsilon)$ neurons can approximate any continuous function on a compact subset of $\mathbb{R}^N$ to within a specified error $\epsilon$. 
In its simplest form, an MLP consists of a finite set of inputs $x_j$, each associated with a learnable weight $\theta_{ij}$, representing the connection strength between input $j$ and neuron $i$, along with a bias term that adjusts the neuron's activation threshold. The output of an MLP can be represented as:

\rev{
\begin{equation}
f(\mathbf{x}) = \sum_{i_L} w_{i_L}^{(L)} \sigma_{i_L}^{(L-1)} + b^{(L)}
\end{equation}
\noindent where, each $\sigma_i^{(\ell)}$ is recursively defined as:
}

\rev{
\begin{equation*}
\sigma_i^{(\ell)} = \begin{cases}
x_i, & \ell = 0 \notag\\
\phi^{(\ell)} \left( \sum_{j} w_{ij}^{(\ell)} \sigma_j^{(\ell-1)} + b_i^{(\ell)} \right), & \ell = 1, \ldots, L-1
\end{cases}
\end{equation*}
}
\rev{
\noindent where, $L$ is the number of layers and $\phi(.)$ denotes the activation function, $w_{ij}$ and $b_{i}$ are the weights and biases respectively. }
The activation function must be a simple, nonlinear, and differentiable function to enable the network to capture complex patterns and compute gradients efficiently during backpropagation. 
Ideally, the activation function is smooth enough to ensure stable gradient behavior during optimization and appropriately bounded to mitigate gradients while maintaining non-saturating derivatives to avoid vanishing gradients. 
\rev{For PDE solving, the activation function must satisfy additional constraints beyond typical machine learning applications: it requires well-defined higher-order derivatives for accurate physics loss computation and sufficient smoothness to ensure stable gradient propagation through the PDE operators.
}

\subsection{Kolmogorov-Arnold Approximation}
\label{sec:kan_rw}

\begin{algorithm}[t]
\caption{A KAN model with a general basis function.}
\begin{algorithmic}[1]
    \Require \texttt{x}: input data of shape \texttt{[batch\_size, l\textsubscript{0}]}, \texttt{network}: \texttt{[l\textsubscript{0}, l\textsubscript{1}, $\ldots$, l\textsubscript{n}]}, \texttt{d}: degree of basis function

    \Ensure \texttt{y}: output of the model for input \texttt{x}

    \State Initialize an empty list \texttt{layers}

    \For{each layer \texttt{(l\textsubscript{i}, l\textsubscript{i+1})} in  \texttt{network} }
        \State Create a new KAN layer \texttt{L\textsubscript{i}} with input dimension          \texttt{l\textsubscript{i}}, output dimension \texttt{l\textsubscript{i+1}}
    
        \State Initialize learnable coefficients \texttt{coeffs\textsubscript{ij}} for each layer \texttt{L\textsubscript{i}}
        \State Append layer \texttt{L\textsubscript{i}} to \texttt{layers}
    \EndFor

    \State \textbf{Forward Pass:}

    \State If required, normalize \texttt{x} to the domain of the basis function

    \For{each layer \texttt{L\textsubscript{i}} in \texttt{layers}}
    
        \State Compute \texttt{B(x)} $\leftarrow$ \texttt{basis\_function(x, d)}
        \State Compute \texttt{y} $\leftarrow \sum_{j=0}^{d}$ \texttt{coeffs\textsubscript{ij}} $\cdot$ \texttt{B\textsubscript{j}(x) }
    \EndFor
    \Return \texttt{y}

    \State \textbf{Function:} \texttt{basis\_function(x, d)}
    \State \quad Generate basis function \texttt{B(x)} from input \texttt{x} and degree \texttt{d}
    \State \quad \textbf{return} \texttt{B(x)}

\end{algorithmic}
\label{algo:kan1}
\end{algorithm}

\rev{
We compare MLP with KAN.} Fig.~\ref{fig:mlp-vs-kan}(middle) illustrates the structure of KAN with learnable activation functions. 
According to Kolmogorov-Arnold representation theorem \cite{kolmogorov1957representation}, any multivariate continuous function $f$ of $n$ variables on a bounded domain can be represented by a superposition of a finite number of continuous functions with a single variable and the summation operation. 
More formally, for any continuous function $f: \mathbb{R}^n \to \mathbb{R}$, there exist continuous functions $ \phi_j $ and $ \psi $ such that:
\begin{equation}
f(\mathbf{x}) = \sum_{i=1}^{2n+1} \psi_i \left( \sum_{j=1}^n \phi_{ij}(x_j) \right)
\end{equation}

\noindent where, $\mathbf{x}$ is the input variable, $\psi_i:\mathbb{R} \rightarrow \mathbb{R}$ and $\phi_{ij}:[0,1] \rightarrow \mathbb{R}$ are parameterized as continuous learnable univariate functions.
\rev{
Liu et al.~\cite{liu2024kan} extend this formula to multiple layers such that: 
}

\rev{
\begin{align*}
\mathbf{x}^{(0)} &= \mathbf{x} \notag\\
x_i^{(\ell)} &= \begin{cases}
\sum_{j=1}^{n_{\ell-1}} \phi_{ij}^{(\ell)}(x_j^{(\ell-1)}), & \ell = 1, \ldots, L-1 \notag\\
\sum_{j=1}^{n_{L-1}} \psi_{ij}(x_j^{(L-1)}), & \ell = L
\end{cases}
\end{align*}
}

The network structure in KANs learns a combination of simpler functions akin to the ensemble methods discussed in Section~\ref{sec:related_works}. 
Specifically,  the base learners $(g_i (x))$ and the weighted sum $\alpha_i$ in the ensemble methods correspond to the inner univariate functions $\phi_{ij}$ and the outer functions $\psi_{ij}$ in KANs, respectively. 
The key difference between KAN and ensemble functions, such as Adaptive Blending Units (ABUs)~\cite{sutfeld2020adaptive},  lies in their approach to function representation. While KAN focuses on building a universal approximation by decomposing a multivariate function into a sum of univariate functions using the Kolmogorov-Arnold representation theorem, the ensemble methods blend multiple activation functions, optimizing their combinations through learned weights to adaptively change the activation shape during training. However, the Kolmogorov-Arnold representation theorem guarantees representation for any continuous function but provides no guarantees about learnability in finite-parameter, gradient-based settings.

In their original work, Liu et al.~\cite{liu2024kan} used the SiLU  activation function to capture the non-linear relationships that could not be captured by the B-spline network as follows:

\begin{equation}
    \phi(x) =\lambda_0 \text{silu}(x) + \lambda_1\text{spline}(x) = \lambda_0 \text{silu}(x) + \lambda_1\sum_{i=1}^{d+k-1} c_i B_i^d(x)
    \label{eq:bspline_silu}
\end{equation}

\noindent where, $d$ is the spline order,  $k$ is the grid size, $\lambda_0$, $\lambda_1$  and $c_i$ are trainable parameters, and $\mathbf{B_i}^d(x)$ is the B-spline function.

Although B-splines~\cite{de1972calculating} are commonly used due to their flexibility and smoothness properties, alternative functional forms such as Chebyshev polynomials~\cite{ss2024chebyshev}, Radial Basis Functions (RBF)~\cite{li2024kolmogorov,ta2024bsrbf}, Fourier~\cite{xu2024fourierkan}, wavelet~\cite{bozorgasl2024wav,seydi2024unveiling}, etc. can also be used for the inner function $\phi$. These alternatives may offer specialized basis functions based on the nature and application of the problem, making KAN a general framework for function approximation. More generally, Algorithm~\ref{algo:kan1} shows the steps for parametrization of the activation function within the KAN formulation. The main idea is to map the inputs of each neural network layer to a higher-dimensional space using a suitable basis function, multiply the result by learnable parameters, and optimize iteratively.

In this study, we examined several well-known basis functions, as shown in Table~\ref{table:basis_functions}, for solving PDEs. The selection criteria are based on their unique characteristics and mathematical properties. For instance, while cubic or high-degree B-splines are differentiable and smooth, the Fourier basis functions have their differentiability tied directly to their coefficients, making them versatile for approximating periodic functions. 

On the other hand, learnable activations can also introduce challenges, particularly in training stability. The increased complexity may lead to issues such as vanishing or exploding gradients during backpropagation, especially in deep networks or when higher-degree polynomials are used. Furthermore, numerical stability may become a concern, as many of these functions are sensitive to inputs outside their well-defined intervals, which can potentially lead to significant calculation errors. 

The Tanh activation function is well-known for suffering from vanishing and exploding gradients when inputs deviate significantly from zero, as its gradient diminishes toward zero when tanh(x)$\rightarrow \pm 1$. Similarly, polynomial basis functions (such as Chebyshev and Jacobi polynomials) can lead to gradient issues. Near-zero polynomial values often cause vanishing gradients, while higher-degree polynomials may cause exploding gradients during backpropagation. This sensitivity to input scale and polynomial degree can destabilize the training process.

Radial Basis Functions (RBFs) are also susceptible to vanishing gradients when the input is far from the center 
$c_i$, especially if the scaling factor $\sigma$ is not optimal, as the gradient approaches zero in these regions. While higher-degree splines generally offer stability, improper scaling or poor placement of knots can induce both vanishing and exploding gradients.
Additionally, Fourier basis functions can cause gradient explosions when high frequencies dominate, resulting in rapid oscillations. This oscillatory behavior significantly increases the gradient magnitude, which can make training unstable, particularly in the presence of high-frequency terms.

These challenges highlight the importance of carefully selecting and scaling hyperparameters and activation functions to maintain stable gradients and prevent issues during backpropagation.

\begin{table}[h!]
\centering
\tiny
\renewcommand{\arraystretch}{1.5}

\caption{KAN basis functions considered in the study and Tanh activation function where,
$O$: the number of parameters in terms of asymptotic notation, 
$n$: the number of units per hidden layer, 
$L$: the number of hidden layers, 
$k$: the grid size, and 
$d$: the degree of the polynomial basis function.}
\label{table:basis_functions}

\begin{tabular}{@{}
                p{0.06\textwidth}
                p{0.15\textwidth}
                ccccc
                p{0.06\textwidth}
                p{0.07\textwidth}
                p{0.20\textwidth}@{}}
\hline \noalign{\vskip 1ex} 

\textbf{Basis Function} & \textbf{Equation} & \textbf{\makecell{Local \\Support}} & \textbf{\makecell{Global \\Support}} & \textbf{\makecell{Node \\Requirement}} & \textbf{\makecell{Numerical \\stability}} & \textbf{\makecell{Smooth\\ness}} & \textbf{Boundedness} & \textbf{\makecell{Number of\\ Parameters}} & \textbf{Suitability} \\ 

\hline \noalign{\vskip 1ex}

B-spline & $B_{i,n}(x) = \frac{x - t_i}{t_{i+n} - t_i} B_{i,n-1}(x) + \frac{t_{i+n+1} - x}{t_{i+n+1} - t_{i+1}} B_{i+1,n-1}(x)$ & \ding{51} & \ding{55} & \ding{51} & \ding{51} & \ding{51} & Locally compact
support & $\mathcal{O}(n^2 \cdot L \cdot (k + d-1))$ & Piecewise Polynomial Approximation, Curve Fitting \\ 

\noalign{\vskip 2ex} 

Radial Basis Function & $e^{-\frac{\|x - c\|^2}{\sigma^2}}$ & \ding{51} & \ding{55} & \ding{51} & \ding{51} & \ding{51} & $(0, 1]$ & $\mathcal{O}(n^2 \cdot L + k)$ & Interpolation, Spatial Modelling \\

\noalign{\vskip 2ex} 

Fourier & $\sum_{k} a_k \cos(kx) + b_k \sin(kx)$ & \ding{55} & \ding{51} & \ding{55} & \ding{51} & \ding{51} & $[-1,1]$ per basis & $\mathcal{O}(n^2 \cdot L\cdot k)$ & Periodic Functions, Signal Processing \\ 

\noalign{\vskip 2ex} 

Chebyshev & $T_d(x) = 2xT_{d-1}(x) +T_{d-2}(x) $ & \ding{55} & \ding{51} & \ding{51} & \ding{51} & \ding{51} & $[-1, 1]$ & $\mathcal{O}(n^2 \cdot L \cdot (d+1))$ & Spectral Methods, Polynomial Approximation \\ 

\noalign{\vskip 2ex} 

Jacobi & $P_d^{(\alpha,\beta)}(x) = \frac{(-1)^d}{2^d d!} (1-x)^{-\alpha} (1+x)^{-\beta} \mathcal{D}^{d} \left[(1-x)^{\alpha+d} (1+x)^{\beta+d}\right]$ & \ding{55} & \ding{51} & \ding{51} & \ding{51} & \ding{51} & $[-1 , 1]$ & $\mathcal{O}(n^2 \cdot L \cdot (d+1))$ & Weighted Polynomial Approximation \\ 

\noalign{\vskip 1ex} \hline \noalign{\vskip 1ex} 

Tanh & $\tanh(x) = \frac{e^x - e^{-x}}{e^x + e^{-x}}$ & \ding{55} & \ding{51} & \ding{55} & \ding{51} & \ding{51} & $[-1, 1]$ & $\mathcal{O}(n^2 \cdot L)$ & Smooth, global pattern \\ 

\noalign{\vskip 1ex} \hline

\end{tabular}

\end{table}

\subsection{Physics Informed Neural Networks}

We integrate PINNs with both MLPs and KANs to directly incorporate physical laws into the network's learning process by embedding governing equations into the loss function.
Fig.~\ref{fig:mlp-vs-kan}(right) shows the structure of a PINN, where the outputs of the network (either MLP or KAN) are used to solve PDEs by enforcing physics-based constraints through automatic differentiation.
The idea is to recover the unique solution of the PDE by penalizing the boundary and initial conditions as data-driven supervised learning and choosing finitely many representative points from the interior that satisfy the differential equation. For instance, to approximate the solution of the PDE in Eq.~\ref{eq:general_pde_eq}, we train the PINN model by minimizing the loss function of the form:

\begin{align}\label{eq:pinn_loss}
    \mathcal{L}(\theta) &= arg\underset{\theta}{min}\sum \limits_{k=1}^{n}  \lambda_k \mathcal{L}_k(\theta) \\
    &= arg\underset{\theta}{min}\Big(\lambda_1 \mathcal{L}_1(\mathcal{D}[u(\mathbf{x});\alpha]  - f(\mathbf{x}) )  +  \sum \limits_{k=2}^{n} \lambda_k \mathcal{L}_k(\mathcal{B_k}[u(\mathbf{x})] - g_k(\mathbf{x}))  \Big) \notag
\end{align}

\noindent where, $n$ is the number of loss terms, and $\lambda_k$ is the weighting coefficient of the respective loss term.

No prior knowledge of the actual solution $u(\mathbf{x})$ is required, as minimizing Eq.~\ref{eq:pinn_loss} yields the desired solution. Eq.~\ref{eq:pinn_loss} reduces the multi-objective constraints of the PDE Eq.~\ref{eq:general_pde_eq}s into a single constraint. 
An important observation is that this loss function includes terms with different physical scales, particularly when high-order derivatives are present in the physics loss term.  This variation can significantly impact the quality and stability of the training process. 
\rev{
Therefore, various adaptive methods have been proposed in the literature to balance the loss weighting terms $\lambda_k$ based on different criteria, such as ~\cite{heydari2019softadapt,mcclenny2020self,wang2021understanding,wang2022and,bischof2025multi,anagnostopoulos2024residual, ZHANG2024109428, zhao2024adaptive}, to name a few.
}

\section{Use Cases}
\label{sec:use-cases}

This section outlines the selected 1D and 2D PDE equations, their boundary conditions, and the loss functions used for evaluation. 
\rev{
These equations encompass various problems, such as the Helmholtz and Wave oscillatory problems, the Klein-Gordon equation representing nonlinear waves, Convection-diffusion in mixed-physics systems, and Cavity flow involving complex fluid dynamics.
}

\subsection{1D Wave Equation}

The wave equation is a second-order hyperbolic PDE that models the propagation of waves, such as sound or vibrations, through a medium. We consider the time-dependent 1D wave equation of the form:

\begin{align}\label{eq:1DWave1}
    u_{tt}(t,x) - c^2u_{xx}(t,x) &= 0   \qquad  (t,x) \in \quad  \Omega \\ 
    u(t,x_0)  &= f_1(t,x)  \qquad (t,x)   \quad \text{on} \quad  \Gamma_0 \notag \\
    u(t,x_1)  &= f_2(t,x) \qquad (t,x) \quad  \text{on} \quad  \Gamma_1 \notag \\
    u(0,x)  &= g(t,x)  \qquad (t,x)  \in \quad \Omega \notag \\
    u_t(0,x) &= h(t,x) \qquad (t,x)  \in \quad  \partial \Omega \notag
\end{align}

\noindent where, $c$ is the wave speed. We set $c=2$, $a=0.5$, $f_1 = f_2 = 0$ and exact solution of the form, $u(t,x) = sin(\pi x) cos(c\pi t) + 0.5 sin(2c\pi x) cos(4 c\pi t)$ on the domain $(t,x) \in [0,1] \times [0,1]$. The equation simplifies to:

\begin{align} \label{eq:1DWave}
    u_{tt}(t,x) -4 u_{xx}(t,x) &= 0 \qquad(t,x) \in \Omega =[0,1] \times [0,1] \\ 
    u(t,0) =u(t,1) &= 0  \notag \\
    u(0,x)  &= sin(\pi x) + 0.5 sin(4\pi x)  \qquad x \in [0,1] \notag \\
    u_t(0,x) &= 0 \notag
\end{align}

\noindent The loss function for solving the problem with PINN is:

\begin{align}    \label{eq:1DWave_Loss1}
\mathcal{L}(\theta) &= \underset{\theta}{min} \big( \lambda_1\|\mathcal{L}_{phy}(\theta)\|_{ \Omega}  +   \lambda_2 \|\mathcal{L}_{bc}(\theta)\|_{ \Gamma_1}   +  \lambda_3\|\mathcal{L}_{ic}(\theta)\|_{ \Gamma_0} \big) \\
&=  \underset{\theta}{min} \big( \lambda_1 \mathcal{L}\|\ u_{\theta_{tt}}(t,x) -4  u_{\theta_{xx}}(t,x)\|_{ \Omega}  +   \lambda_2 \mathcal{L}\| u(t,0) + u(t,1) \|_{ \Gamma_1}  +  \lambda_2 \| u(0,x) \|_{ \Gamma_0}   +  \lambda_3 \mathcal{L}\| u_{\theta_{t}}(0,x) \|_{ \partial \Omega} \big) \notag
\end{align}

\noindent We set $\lambda_1 = 1.0$, $\lambda_2 = 100.0$, and $\lambda_3 = 100.0$. The difficulty of this problem lies in the presence of both low- and high-frequency components in the solution, as well as the nonlinearity introduced by the interaction between time and space.


\subsection{Helmholtz Equation}

The Helmholtz equation is a time-independent form of the wave equation arising in many physical problems like vibrating membranes, acoustics, and electromagnetism.
We consider a 2D Helmholtz equation of the form:

\begin{align}\label{eq:Helmholtz1} 
    \Delta u(x,y) +k^2u(x,y) &= q(x,y) \qquad(x,y) \in \Omega \\ 
    u(x,y) &= h(x,y) \qquad (x,y) \in \Gamma_0 \notag
 \end{align}
 
\noindent where, $\Delta$ is the Laplacian operator ($\Delta u = u_{xx} + u_{yy}$ ), $q(x,y)$ is the forcing term , and $k$ is the wavenumber. For simplicity, we assume Dirichlet boundary conditions ($h(x,y)$) equal to an exact solution of the form, $u(x,y) = sin(a_1  \pi x)  sin(a_2  \pi y)$ corresponding to the source term $q(x,y) = u(x,y)[k^2 - (a_1\pi)^2 -(a_2\pi)^2]$ with $k = 1$, $a_1 = 1$, and $a_2 = 4$. The final form of the equation with the boundary and initial conditions is: 

\begin{align}\label{eq:Helmholtz} 
    \Delta u(x,y) + u(x,y)[ (a_1\pi)^2+(a_2\pi)^2] &= 0 \qquad(x,y) \in \Omega = [-1,1]\times[-1,1]\\ 
        u(x,y) &= 0 \qquad (x,y) \in \Gamma_0 \notag
\end{align}

\noindent The loss function for solving the problem with PINN is:

\begin{align}\label{eq:Helmholtz_loss}
    \mathcal{L}(\theta) &= \underset{\theta}{min} \big( \lambda_1\|\mathcal{L}_{phy}(\theta)\|_{ \Omega}  +   \lambda_2 \|\mathcal{L}_{bc}(\theta)\|_{ \Gamma_0}  \big) \\
    &=  \underset{\theta}{min} \big( \lambda_1 \mathcal{L}\|\ u_{\theta_{xx}}(x,y) +  u_{\theta_{yy}}(x,y) + \alpha u_{\theta}(x,y) \|_{ \Omega}  +   \lambda_2 \mathcal{L}\| u_{\theta}(x,y) - u(x,y) \|_{ \Gamma_0}  \big) \notag
\end{align}

We set $\lambda_1 = 1.0$, and $\lambda_2 = 10.0$.

\subsection{Klein-Gordon Equation}

Klein-Gordon equation is a second-order hyperbolic PDE that appears in various fields such as quantum field theory, general relativity, and nonlinear optics. We consider 1D nonlinear Klein-Gordon equation of the form:
\begin{align}\label{eq:Klein-Gordon1}
        u_{tt} - \alpha u_{xx} + \beta u +  \gamma u^k &= 0 \qquad(t ,x) \in \Omega\\ 
        u(t,x) &= g_1(t,x) \qquad (t,x) \notag\\
        u_t(t,x) &= g_2(t,x)\qquad (t,x) \notag\\
        u(0,x) &= h(t,x)  \qquad (t,x) \in \partial \Omega \times [0,T] \notag
\end{align}

\noindent where, $\alpha$, $\beta$, $\gamma$ and $k$ are known constants. We set $1,0,1,3$ for $\alpha$, $\beta$, $\gamma$ and $k$ respectively. We use the exact solution of the form, $u(t,x) = xcos (5 \pi t) +(tx)^3 $. The final form of the equation with the boundary and initial conditions is:

\begin{align}\label{eq:Klein-Gordon}
        u_{tt}(t,x)- \alpha u_{xx}(t,x) + \beta u(t,x) +  u^3(t,x) &= 0 \qquad(t ,x) \in \Omega =[0,1] \times [0,1] \\ 
        u(t,0) &= 0  \notag\\
        u(t,1) &=  cos (5 \pi t) +(t)^3  \notag\\
        u(0,x) &=x  \notag\\
        u_t(0,x) &=0  \notag
\end{align}

\noindent The  loss function for solving the problem with PINN is:

\begin{align}    \label{eq:Klein-Gordon_loss}
\mathcal{L}(\theta) &= \underset{\theta}{min} \big( \lambda_1\|\mathcal{L}_{phy}(\theta)\|_{ \Omega}  +   \lambda_2 \|\mathcal{L}_{bc}(\theta)\|_{ \Gamma_1}   +  \lambda_3\|\mathcal{L}_{ic}(\theta)\|_{ \Gamma_0} \big)  \\
&=  \underset{\theta}{min} \big( \lambda_1 \mathcal{L}\|\  u_{\theta_{tt}}(t,x) - u_{\theta_{xx}}(t,x) + u^3(t,x) \|_{ \Omega}  +   \lambda_2 \mathcal{L}\| u(t,0) + u(t,1) + u(0,x) \|_{ \Gamma_1}   +  \lambda_3 \mathcal{L}\| u_{\theta_{t}}(0,x) \|_{ \partial \Omega} \big) \notag
\end{align}

\noindent We set $\lambda_1 = 1.0$ , $\lambda_2 = 50.0$, and $\lambda_3 = 50.0$.

\subsection{Convection-diffusion Equation}

The convection-diffusion equation is a second-order PDE that models the transport of a substance, such as heat, in a medium, taking into account both convection (movement due to flow) and diffusion (spreading out due to concentration gradients). We consider the viscous 2D convection-diffusion equation of the form:
\begin{align}\label{eq:Convection-Diffusion1}
        u_{t} + c_1 u_{x} + c_2 u_{y} - D \Delta u(x,y) &= 0 \qquad (t, x, y) \in \Omega \\
        u(t, \mathbf{x}) &= g_0(t, \mathbf{x}) \qquad (t, \mathbf{x}) \in \Gamma_0 \notag\\
        u(t, \mathbf{x}) &= g_1(t, \mathbf{x}) \qquad (t, \mathbf{x}) \in \Gamma_1 \notag\\
        u(0, \mathbf{x}) &= h(0 , \mathbf{x}) \qquad \mathbf{x} \in \partial \Omega \times [0,T] \notag
\end{align}

\noindent where,  $c_1$ and $c_2$ are the convection velocities in the $x$ and $y$ directions respectively, $D$ is the diffusion coefficient, and $\Delta$ is the Laplacian operator. We use an exact solution of the form $u(t, x, y) = \exp(-100((x-0.5)^2 + (y-0.5)^2)) \exp(-t)$ which describes a Gaussian pulse centered at $(x,y)=(0.5,0.5) $ that decays over time. The corresponding to the initial condition is $h(0 ,x, y) = \exp(-100((x-0.5)^2 + (y-0.5)^2))$. For this particular problem, later timesteps are important because the system has evolved, and the solution becomes more diffuse and less concentrated. As a result, it becomes harder for models to accurately capture the subtle diffusion and dissipation effects.

We set $c_1 = 1.0$, $c_2 = 1.0$ (high convection), and $D = 0.01$ (low diffusion). The final form of the equation with the boundary and initial conditions is:
\begin{align}\label{eq:Convection-Diffusion}
    u_{t}(t, x, y) &+ c_1 u_{x}(t, x, y) + c-2 u_{y}(t, x, y) \\
    &- D \left( u_{xx}(t, x, y) + u_{yy}(t, x, y) \right) = 0 \quad (t, x, y) \in \Omega = [0,1] \times [0,1] \times [0,1] \notag \\
    u(t, x, y) &= g(t, x, y) \qquad (t, x, y) \in \Omega \notag\\
    u(0, x, y) &= h(x, y) \qquad (x, y) \in \partial \Omega \times [0,T] \notag
\end{align}

\noindent The loss function for solving the problem with PINN is:

\begin{align}\label{eq:Convection-Diffusion_loss}
    \mathcal{L}(\theta) &= \underset{\theta}{\text{min}} \big( \lambda_1 \|\mathcal{L}_{\text{phy}}(\theta)\|_{\Omega} + \lambda_2 \|\mathcal{L}_{\text{bc}}(\theta)\|_{\Gamma_1} + \lambda_3 \|\mathcal{L}_{\text{ic}}(\theta)\|_{\Gamma_0} \big) 
    \\
    &= \underset{\theta}{\text{min}} \big( \lambda_1 \|\ u_{\theta_{t}}(t, x, y) + v_x u_{\theta_{x}}(t, x, y) + v_y u_{\theta_{y}}(t, x, y) - D (u_{\theta_{xx}}(t, x, y) + u_{\theta_{yy}}(t, x, y)) \|_{\Omega} \notag
    \\
    &\quad \quad + \lambda_2 \mathcal{L}\| u(t, x, y) - g_1(t, x, y) \|_{\Gamma_1} + \lambda_3 \mathcal{L}\| u_{\theta}(0, x, y) - h(x, y) \|_{\Omega_0} \big) \notag
\end{align}

\noindent We set $\lambda_1 = 1.0$, $\lambda_2 = 10.0$, and $\lambda_3 = 10.0$.

\subsection{Time-dependent 2D Cavity Problem}
This problem is a widely recognized benchmark in Computational Fluid Dynamics (CFD). It involves simulating the movement of cavity walls to drive fluid flow within the cavity. In this case study, the upper boundary moves at a constant velocity, creating a shear-driven flow inside the cavity, while the other boundaries remain stationary. The problem is governed by the unsteady, incompressible Navier-Stokes equations as follows:

\begin{align}     \label{eq:Cavity}
    \rho \left(\frac{\partial \mathbf{u}}{\partial t} + \mathbf{u} \cdot \nabla \mathbf{u}\right) &= - \nabla p + \mu \nabla^2 \mathbf{u}  \qquad \text{in}  \quad \Omega \in [0 , 1]\\
    \nabla \cdot \mathbf{u} &= 0  \quad \text{in} \quad \Omega \notag\\
    \mathbf{u}(0 , x) &= 0  \qquad \text{in} \quad \Omega  \notag\\
    \mathbf{u}(t , x_0) &= 0 \qquad \text{on} \quad \Gamma_1  \notag\\
    \mathbf{u}(t , x_l) &= 1 \qquad \text{on} \quad \Gamma_0  \notag
\end{align}

The computation domain  $\Omega$ is a two-dimensional square cavity $\Omega$ $ = (0, 1) \times (0, 1)$.  Uniform discretization is used with grid $(N_x , N_y) = (100 , 100)$ and 10 seconds total simulation time with 0.01s time interval. $\Gamma_1$ is the top Dirichlet boundary condition of the cavity with velocity tangent to this side, $\Gamma_0$ denotes the other three stationary sides,  $\rho $= 1056, and $ \mu = \frac{1}{Re} = 0.01$; where, $Re$ is Reynolds number of the flow and $\rho$ is fluid density. 
\rev{We used Ansys Fluent software with the finite volume method to generate the reference solution and compare the results with the neural network approximation}.
The loss function for solving the Cavity problem is as follows:

\begin{align}    \label{cavity_loss}
    \mathcal{L}(\theta) &= \lambda_1|\mathcal{L}_{phy}\|_{ \Omega}  +    \lambda_2  \big(\|\mathcal{L}_{up}\|_{ \Gamma_1}   +   \|\mathcal{L}_{bc_{1}}\|_{ \Gamma_0} \big) +  \lambda_3 \|\mathcal{L}_{u0}\|_{\Omega}  
\end{align}

\noindent where, $ \mathcal{L}_{phy}(\theta)=  \mathcal{L}_{r_u} +  \mathcal{L}_{r_v} + \mathcal{L}_{r_c}$. We select $\lambda_1 = 0.1$, $\lambda_2 = 2.0$, and $\lambda_3 = 4.0$ such that:

\begin{align*}
    \mathcal{L}_{r_u}(\theta) &= \text{MSE} \left[(\hat{u}_t + \hat{u}  \hat{u}_x + \hat{v} \hat{u}_y) + \frac{1.0}{ \rho}  \hat{p}_x - \mu  (\hat{u}_{xx} + \hat{u}_{yy})\right]\\
    \mathcal{L}_{r_v}(\theta) &=\text{MSE}\left[(\hat{v}_t + (\hat{u}  \hat{v}_x + \hat{v}  \hat{v}_y+ \frac{1.0}{ \rho}  \hat{p}_y - \mu  (\hat{v}_{xx}+ \hat{v}_{yy})\right]\\
    \mathcal{L}_{r_c}(\theta) &= \text{MSE}\left[(\hat{u}_x + \hat{v}_y)\right]
\end{align*}
    \noindent with the boundary and initial conditions:
\begin{align*}
 \mathcal{L}_{up} &=  \text{MSE}\left[(1.0 - \hat{u} + (\hat{v})\right]\\ 
    \mathcal{L}_{bc_{1}}& = \mathcal{L}_{\text{bottom, right, left}} =  \text{MSE}\left[( \hat{u} + \hat{v}\right]\\ 
    \mathcal{L}_{u_{0}} &= \text{MSE}\left[ \hat{u}) + (\hat{v})+  (\hat{p}\right]\\ 
\end{align*}

\begin{table}[t]
\scriptsize
\centering
\caption{\rev{KAN learnable activation function with the hyperparameters used in the study where $a$, $b$, $a_{ijr}$, $b_{ijr}$, and $c_{ijr}$ are learnable parameters.}}
\begin{tabular} {
    @{}
    l@{\hspace{17.0pt}}
    l@{\hspace{10.0pt}}
    l@{\hspace{17.0pt}}
    @{}
}
    \toprule
    \textbf{Type} & \textbf{Hyperparameters} & \textbf{Activation Function} $f(z)$ \\
    \midrule
    Tanh & - & $f(\mathbf{z}) = \tanh\left( \sum_{j=1}^{n_{l}} w_{ij} z_j + b_i \right)$  \\
    \hline
    \makecell{parametric-Tanh} & - & $f(\mathbf{z}) = \tanh\left( a \cdot \left( \sum_{j=1}^{n_{l}} w_{ij} z_j + b_i \right) +b\right)$\\
    \hline
    B-spline & \makecell[l]{$g=8$,\\ $d=3$} & 
    \makecell[l]{
    $f(\mathbf{z}) = \sum_{i=1}^{n_{l}} \left( \sum_{j=1}^{n_{l-1}} \sum_{r=1}^{g+d+1} c_{ijr} \cdot B_r^3(z_j) \right)$ where: \\
    $\quad B_r^3(z) = \begin{cases}
    \frac{1}{6}(z - t_r)^3 & \text{if } z \in [t_r, t_{r+1}) \\
    \frac{1}{6}((t_{r+1} - z)^3 - 4(z - t_{r+1})^3) & \text{if } z \in [t_{r+1}, t_{r+2}) \\
    \frac{1}{6}((t_{r+3} - z)^3 - 4(t_{r+2} - z)^3) & \text{if } z \in [t_{r+2}, t_{r+3}) \\
    \frac{1}{6}(t_{r+4} - z)^3 & \text{if } z \in [t_{r+3}, t_{r+4}) \\
    0 & \text{otherwise}
    \end{cases}$ 
    }\\
    \hline
    GRBF & \makecell[l]{$\sigma=1$,\\ $k=8$} & $f(\mathbf{z}) = \sum_{i=1}^{n_l} \left( \sum_{j=1}^{n_{l-1}} \sum_{r=1}^{k} c_{ijr} \cdot e^{-\frac{\|z_j - t_r\|^2}{\sigma^2}} \right)$ \\
    \hline
    Fourier & $k=4$ & $f(\mathbf{z}) = \sum_{i=1}^{n_l} \left( \sum_{j=1}^{n_{l}} \sum_{r=1}^{k} \left[ a_{ijr} \cos(k z_j) + b_{ijr} \sin(k z_j) \right] \right)$\\
    \hline
    Chebyshev & $d=4$ & 
    \makecell[l]{
    $f(\mathbf{z}) = \sum_{i=1}^{n_l} \left( \sum_{j=1}^{n_{l-1}} \sum_{r=0}^{d} c_{ijr} \cdot T_r(z_j) \right)$ where:\\
        $\quad T_0(z) = 1$, \quad  $T_1(z) = z$,\\
        $\quad T_2(z) = 2z^2 - 1$,\\
        $\quad T_3(z) = 4z^3 - 3z$,\\
        $\quad T_4(z) = 8z^4 - 8z^2 + 1$ }\\
    \hline
    Jacobi & 
    \makecell[l]{$d=4$,\\$\alpha=1$,\\ $\beta=1$ }& 
    \makecell[l]{
    $f(\mathbf{z}) = \sum_{i=1}^{n_l} \left( \sum_{j=1}^{n_{l-1}} \sum_{r=0}^{d} c_{ijr} \cdot P_r^{(1,1)}(z_j) \right)$  where:\\
        $\quad P_0^{(1,1)}(z) = 1$, \quad $P_1^{(1,1)}(z) = 2z$, \\
        $\quad P_2^{(1,1)}(z) = \frac{3}{2}(5z^2 - 1)$, \\
        $\quad P_3^{(1,1)}(z) = \frac{5}{2}(7z^3 - 3z)$, \\
        $\quad P_4^{(1,1)}(z) = \frac{35}{8}(9z^4 - 10z^2 + 1)$ }\\
    \bottomrule 

\end{tabular}
\label{table:model_formulations}
\end{table}

\begin{table}[t]
\centering
\scriptsize
\caption{
Two different neural network architectures (A1 and A2) and their corresponding parameters for MLP and KAN models. 
Architectures A1 and A2 differ in the number of units (neurons) per layer and the overall complexity of the network, 
\rev{with A1 being smaller and A2 being larger, making A2 more expressive.} Cells marked with ``$-$'' indicate that the particular configurations are not used.
}

\begin{tabular}{@{}ll
        l@{\hspace{4.0pt}}
        l@{\hspace{4.0pt}}
        l@{\hspace{4.0pt}}@{\hspace{4.0pt}} |
        l@{\hspace{4.0pt}}
        l@{\hspace{4.0pt}}
        l@{\hspace{4.0pt}}
        l@{\hspace{4.0pt}}
        l@{\hspace{4.0pt}}
        @{}}
    \toprule
    
    \multirow{2}{*}{\textbf{Arch}} & \multirow{2}{*}{\textbf{Case}} &  \multirow{2}{*}{\textbf{Network}} &\multicolumn{2}{c|}{\textbf{MLP}} &\multicolumn{5}{c}{\textbf{KAN}} \\

    \cline{4-10} \noalign{\vskip 1ex}       
    
    & &   &\makecell{\textbf{Tanh}} & \makecell{\textbf{Param-Tanh}} &\makecell{\textbf{B-spline}} &\makecell{\textbf{GRBF}} &\makecell{\textbf{Fourier}} &\makecell{\textbf{Chebyshev}} &\makecell{\textbf{Jacobi}}\\
        
    \midrule
        
    & Helmholtz &  $[2, 30, 30, 30, 1]$ &  1981& 1983  & 18900  & 15120  & 15211  &9450  &9450  \\   
    & {\makecell[l]{Klein-Gordon}}& $ [2, 30, 30, 30, 1]$   & 1981  &1983  &18900 & 15120  & 15211 & 9450 & 9450  \\
    
    \textbf{A1} & Wave  &   $[2, 10, 30, 10, 1]$   &681  &683 &6300  &5040  & 5091  & 3150  & 3150  \\  
    
    & {\makecell[l]{Convection-Diffusion}}  &$[3, 50, 50, 50, 1]$   & 5351  & 5353  & 52000  &41600  & 41751  & 26000  & 26000 \\                              
    & Cavity   & $[3, 50, 50, 50, 3]$  & 5453  & 5455  &53000 & 42400  & 42553  &  26500 & 26500  \\
    
    \midrule
        
    & Helmholtz &  $[2, 300, 300, 300, 1]$  &181801 &181803 & -  & -  & -  & -  & -  \\   
    & {\makecell[l]{Klein-Gordon}}& $[2, 400, 400, 400, 1]$   &322401  & 322403  & -  & -  & -  & -  & -  \\
    
    & Wave  &  $ [2, 300, 300, 300, 300, 1]$   &272101  & 272103  & -  & -  & -  & -  & -  \\  
    
    & {\makecell[l]{Convection-Diffusion}}  & $[3, 300, 300, 300, 1]$   &  182101 & 182103  & -  & -  & -  &-  &-  \\                              
   \multirow{2}{*}{\makecell[l]{\textbf{A2}}}  & Cavity   &  $[3, 300, 300, 300, 3]$  & 182703  &182705  & -  &-  & -  &- &-  \\

    \cline{2-10} \noalign{\vskip 1ex}       
        
    & Helmholtz &  $[2, 150, 150, 150, 1]$  &- &- & 454500 & 363600  & 364051  & 227250  & 227250  \\   
    & {\makecell[l]{Klein-Gordon}}  & $[2, 200, 200, 200, 1]$   & -  & -  & 806000  &644800  & 645401  & 403000  & 403000 \\
    
    & Wave &  $ [2, 150, 150, 150, 150, 1]$   & - &- & 679500  &543600  & 544201  & 339750  & 339750  \\  
    
    & {\makecell[l]{Convection-Diffusion}}& $[3, 150, 150, 150, 1]$   &-  & -  & 456000 &364800  & 365251  & 228000  & 228000  \\                              
    &  Cavity &  $[3, 100, 100, 100, 3]$  & -  &-  & 206000 & 164800  & 165103  & 103000  &103000  \\
    
    \bottomrule
    
\end{tabular}
\label{table:network_architecture}
\end{table}

\begin{table}[t]
    \centering
    \tiny
    \caption{
    Testing relative $L_2$ error (in \%) for the cases considered in our study. 
    We used two neural network architecture sets (A1 and A2), detailed in Table~\ref{table:network_architecture}. 
    The \green{lowest} and the \blue{second-lowest} errors are highlighted. 
    For simplicity, we select the best model based on the accuracy of the first feature in each row, which corresponds to the velocity $u$. Cells marked with ``$-$'' indicate that the error exceeded 90\%. $t/it$: time per iteration in seconds.}
    \label{table:results}
        \begin{tabular}{@{}ll
        c!{\color{lightgray}\vline}
        c!{\color{lightgray}\vline}
        c!{\color{lightgray}\vline}
        c!{\color{lightgray}\vline}
        c!{\color{lightgray}\vline}
        c!{\color{lightgray}\vline}
        c!{\color{lightgray}\vline}
        c!{\color{lightgray}\vline}
        c!{\color{lightgray}\vline}
        c!{\color{lightgray}\vline}
        c!{\color{lightgray}\vline}
        c!{\color{lightgray}\vline}
        c!{\color{lightgray}\vline}
        c
        c@{}}
    
        \toprule
        
        \multirow{3}{*}{\textbf{Case}} & \multirow{3}{*}{} & \multicolumn{4}{c|}{\textbf{MLP}} & \multicolumn{9}{c}{\textbf{KAN}} \\
        
        \cline{3-7} \cline{8-16} \noalign{\vskip 1ex}
          
         & & 
         \multicolumn{2}{c|}{\textbf{Tanh}} & 
         \multicolumn{2}{c|}{\textbf{param-Tanh}}  & 
         \multicolumn{2}{c|}{\textbf{B-spline}} & 
         \multicolumn{2}{c|}{\textbf{GRBF}} & 
         \multicolumn{2}{c|}{\textbf{Fourier}} & 
         \multicolumn{2}{c|}{\textbf{Chebyshev}} & 
         \multicolumn{2}{c}{\textbf{Jacobi}}  \\
         
        \cline{3-7} \cline{8-16} \noalign{\vskip 1ex} 
         && A1 & A2 & A1 & A2 & A1 & A2 & A1 & A2 &  A1 & A2 &  A1& A2& A1 & A2 \\
        
        \midrule
        
        & $u$ & 6.13 & 6.62 &  5.51 & 4.52 & \greencell{1.93} & 3.03 & 17.9 & 8.11 & \bluecell{3.09} & $-$ & 5.17 & $-$ & 19.5 &$-$ \\  
          
Helmholtz & $f$ & 3.09 & 1.63  & 2.50 & 0.96 & \greencell{0.92} & 0.71  & 4.52 & 3.76 & \bluecell{1.18} & $-$ & 1.37 & $-$ & 2.01 & $-$ \\   
          & $t/it$& 0.0111 & 0.0116 & 0.0145 &  0.0153  & 0.0904  & 0.0907 &  0.0238 &  0.0243 &  0.0309 &  $-$ & 0.0934 &  $-$ &   0.1376 &   $-$  \\  
        \midrule
        
\multirow{3}{*}{\makecell[l]{Klein-\\Gordon}} & $u$ & \greencell{1.58} & 9.48 & 11.4 & 18.6  & \bluecell{5.23} & 13.2   & 15.0 &  25.8  & 9.25 &  $-$  & 12.7 & $-$ & 8.36 & $-$ \\   

        & $f$ & \greencell{5.08} & 4.85 & 2.22 & 1.77 & \bluecell{7.89} & 0.33 & 0.22 & 0.43 & 0.62 &  $-$ & 1.07 & $-$  & 0.46 &  $-$  \\   
          & $t/it$& 0.0153 &  0.0131 & 0.0187 & 0.0154 & 0.1028 &   0.0841 &  0.0260 &  0.0254 &  0.0367 & $-$ &  0.1044 &  $-$&  0.1354 & $-$ \\   
        \midrule
        
\multirow{2}{*}{Wave} & $u$ &  $-$ & \greencell{13.6} & $-$ & $-$ &  22.4 & 27.7 & 17.2 & \bluecell{14.7}  &  38.4 & $-$ & 37.5 &  44.2 & 37.5 & $-$\\              
          & $t/it$& $-$ & 0.0146 &   $-$ &  $-$  & 0.0823 &  0.1030 &  0.0217 &  0.0266 &  0.0294 &   $-$ & 0.0889 &  0.1167 &   0.1328 & $-$ \\  
       
    \midrule
       
   \multirow{3}{*}{\makecell[l]{Convection\\Diffusion}} & $u$ &12.6 & 13.6& 12.1 & \bluecell{11.6}& 11.7 &  12.3  & \greencell{11.5} &  12.8  & 15 & $-$ & 17.5 & $-$ & 29.9 & $-$ \\ 
    & $f$ & 2.45 & 3.86  &  1.67 &  \bluecell{1.07}  & 8.34 & 1.01 & \greencell{2.07} & 3.01 & 7.10 & $-$ & 2.46 & $-$ & 3.02 & $-$ \\ 
          & $t/it$&   0.0132 &  0.0170 &   0.0177 &   0.0174 & 0.0975 &  0.0991 & 0.0277 &  0.0299 & 0.0340 & $-$ & 0.1005 & $-$ &   0.1351 & $-$ \\  
        
    \midrule
        
        & $u_x$ & 13.9 & \bluecell{10.8}  & 13.5 & 15.7  & 7.87  &  \greencell{4.09} & 13.1 &  13.5   &25.2 &  40.9 & 17.2 & 14.6   & 16.9 & 37.1 \\   
Cavity  & $v_x$ & 24.3 & \bluecell{22.0}  & 32.8 & 21.9  &  18.0 & \greencell{8.72} &  25.6 & 17.7  &38.9 & 75.2 & 32.5 &  24.2   &44.2 & 65.2 \\   
        & $p$   & 25.9 & \bluecell{24.4}  & 45.8 & 33.5  & 25.4 & \greencell{21.3} &  28.2 & 29.7  & 52.2 &  $-$ &  37.1  & 52.2  & 54.1 & $-$ \\ 
          & $t/it$& 0.0282 & 0.0271 &    0.0333 &  0.0358 & 0.2030 &  0.1982 & 0.0537 &  0.0538 &  0.0771 &  0.0760 &  0.2137 &  0.2255 &  0.2923 &  0.2876 \\   
        \bottomrule
    \end{tabular}
\end{table}

\begin{table}[t]
    \centering
    \tiny
    \caption{
    The final training losses include the physics-informed loss ($L_{phy}$), the boundary conditions loss ($L_{bc}$), and the initial conditions loss ($L_{ic}$). Information about these loss functions can be found in Section~\ref{sec:use-cases}.
    }

    \begin{tabular}{@{}ll
    c@{\hskip 0.08cm}!{\color{lightgray}\vline}
    c@{\hskip 0.08cm}!{\color{lightgray}\vline}
    c@{\hskip 0.08cm}!{\color{lightgray}\vline}
    c@{\hskip 0.08cm}!{\color{lightgray}\vline}
    c@{\hskip 0.08cm}!{\color{lightgray}\vline}
    c@{\hskip 0.08cm}!{\color{lightgray}\vline}
    c@{\hskip 0.08cm}!{\color{lightgray}\vline}
    c@{\hskip 0.08cm}!{\color{lightgray}\vline}
    c@{\hskip 0.08cm}!{\color{lightgray}\vline}
    c@{\hskip 0.08cm}!{\color{lightgray}\vline}
    c@{\hskip 0.08cm}!{\color{lightgray}\vline}
    c@{\hskip 0.08cm}!{\color{lightgray}\vline}
    c@{\hskip 0.08cm}!{\color{lightgray}\vline}
    c@{\hskip 0.08cm}
    c@{}
    }
    
        \toprule
        
        \multirow{3}{*}{\textbf{Case}} & \multirow{3}{*}{} & \multicolumn{4}{c|}{\textbf{MLP}} & \multicolumn{9}{c}{\textbf{KAN}} \\
        
        \cline{3-7} \cline{8-16} \noalign{\vskip 1ex}
          
         & & 
         \multicolumn{2}{c|}{\textbf{Tanh}} & 
         \multicolumn{2}{c|}{\textbf{param-Tanh}}  & 
         \multicolumn{2}{c|}{\textbf{B-spline}} & 
         \multicolumn{2}{c|}{\textbf{GRBF}} & 
         \multicolumn{2}{c|}{\textbf{Fourier}} & 
         \multicolumn{2}{c|}{\textbf{Chebyshev}} & 
         \multicolumn{2}{c}{\textbf{Jacobi}}  \\
         
        \cline{3-7} \cline{8-16} \noalign{\vskip 1ex} 
         && A1 & A2 & A1 & A2 & A1 & A2 & A1 & A2 &  A1 & A2 &  A1& A2& A1 & A2 \\
        
        \midrule

   Helmholtz & $L_{phy}$ & 4.4e+00  & 1.91e+00 &   2.04e+00 & 2.4e-01 & 6.1e-01 & 2.75e-01 &  1.04e+01 & 2.7e+00 & 1.0e+00 & 3.30e+12 & 8.3e-01 & 7.15e+03 & 2.9e+00 & 6.80e+03 \\  
          
        & $L_{bc}$ & 6.6e-03 & 3.27e-03  & 6.55e-03 &  3.1e-03 & 1.5e-03 & 1.66e-03  & 1.30e-02 & 2.1e-02 & 2.7e-03 & 3.39e-01 & 2.8e-03 & 1.13e-01 & 2.3e-02 & 1.50e-01 \\   
        \midrule
        
\multirow{3}{*}{\makecell[l]{Klein-\\Gordon}} &$L_{phy}$  & 1.9e-01 & 8.57e+00  & 1.3e+00 & 1.31e+00 & 3.0e-01 & 2.01e-01 & 1.2e-01 &  2.64e-01 & 4.6e-01 & 2.29e+06 &  2.4e+00 & 1.19e+04 & 3.5e-01 & 9.46e+03 \\      

        & $L_{bc}$  & 6.0e-05 & 3.86e-04  & 7.4e-04 & 7.39e-04 & 4.2e-05 & 1.00e-05  & 2.5e-05&  4.74e-04 & 5.7e-04 & 5.96e-01 & 8.6e-04 & 4.16e-02 & 3.9e-04 & 5.21e-02 \\      
        & $L_{ic}$ & 4.5e-06 & 4.74e-04  & 7.5e-05 &  7.47e-05 & 4.8e-06 & 4.34e-06 & 6.1e-06 & 4.83e-04 & 8.1e-05 & 3.30e-01 & 6.6e-05 & 2.32e-01 & 4.3e-04 &  1.82e-01 \\   
        \midrule
        
\multirow{2}{*}{Wave} & $L_{phy}$ &  1.47e-01 & 1.2e-01  &  3.24e-02 & 3.7e-02 & 3.1e-02 &  3.00e-02  & 8.83e-01 & 6.6e-01 & 1.3e+00& 1.65e+10 & 2.9e-01 & 3.77e-01 & 8.9e-01 & 1.89e-07 \\                 
        & $L_{bc}$  &1.46e-02 & 4.3e-04  & 1.09e-02 & 1.7e-04 & 5.0e-05 & 2.74e-05 &  1.45e-03 & 1.6e-03& 1.3e-02 & 9.30e-02 & 1.5e-02 &  3.50e-02  & 1.8e-02 & 5.97e-02 \\   
        & $L_{ic}$  &  3.54e-03 & 7.8e-05  & 4.59e-03 & 3.6e-05 & 7.6e-06 & 9.23e-06 & 6.64e-04 & 5.8e-04 & 1.3e-02 & 3.92e-01 & 3.2e-03&  1.15e-02 & 7.7e-03& 4.50e-01 \\
    \midrule
       
   \multirow{3}{*}{\makecell[l]{Convection\\Diffusion}} &$L_{phy}$  & 7.2e-04  & 1.74e-03 & 2.74e-04 & 5.8e-03 & 6.0e-05 &  5.30e-05 & 2.7e-04 & 7.28e-03 & 6.3e-03 &  5.52e+05 & 1.1e-03 & 1.02e+00 & 8.5e-04 & 9.69e-01 \\    
    & $L_{bc}$  & 3.0e-05 & 3.16e-05 & 2.15e-05 & 2.6e-05 & 8.3e-06 & 2.28e-05 & 2.3e-05 & 4.39e-05 & 1.3e-04& 4.50e-03 &1.6e-04 & 1.02e-02 &  8.6e-04 & 3.67e-05 \\   
        & $L_{ic}$  &  5.4e-05 & 8.99e-05 &  7.77e-05 & 1.2e-04 & 5.3e-06 & 4.31e-06  &6.3e-05 & 7.21e-05 & 1.3e-04 & 1.37e-02 & 1.2e-04 & 8.87e-03 & 4.2e-04 & 1.61e-02 \\   
    \midrule
        
        & $L_{phy}$  & 1.42e-01 & 2.2e-02  & 1.3e-01  & 2.79e-02 & 2.56e-01 &2.3e-02  & 1.4e-01 & 6.75e-02 & 2.1e-01 & 3.25e-01 & 1.34e-01 & 2.9e-02 & 2.4e-01 & 1.86e-07 \\    
Cavity  & $L_{bc}$ & 1.35e-02 & 4.5e-04  & 3.4e-04 & 1.08e-02 & 2.95e-03 & 1.5e-05  & 4.4e-03 & 5.06e-03 & 3.6e-03 & 1.50e-02 & 1.39e-02 & 1.4e-03 & 9.1e-03 & 1.02e+00 \\     
        & $L_{ic}$  & 2.57e-03 & 6.6e-03  & 4.0e-03  & 3.81e-03 & 6.53e-04 & 2.5e-04  & 5.7e-03& 4.57e-03 & 7.6e-03 & 1.04e-02 & 3.81e-03 & 3.7e-03 & 6.7e-03 &  8.42e-02 \\   
        \bottomrule
    \end{tabular}
    \label{table:final-loss}
\end{table}

\section{Computational Experiments}
\label{sec:computational-experiment}

\subsection{Experiment Setup}

\rev{
Table~\ref{table:model_formulations} shows the hyperparameters for the KAN basis functions, which are explained in Table~\ref{table:basis_functions}.
}
Table~\ref{table:network_architecture} presents two groups of neural network architectures (Fig.~\ref{fig:mlp-vs-kan}) along with their corresponding settings used in this study, as well as the total number of learnable parameters for each model. We designate these architectures as A1 and A2, where A1 is designed with fewer parameters while A2 includes more parameters, thereby increasing model complexity and enhancing expressivity for function approximation.

The settings used in this study are selected to achieve optimal performance for moderately complex problems while minimizing the risk of overfitting and reducing computational complexity. Therefore, we employed identical training settings for shared parameters to ensure a fair comparison of the models. All models were trained using the PyTorch Adam optimizer with hyperparameters $\beta_1 = 0.9$, $\beta_2 = 0.999$, $\epsilon = 10^{-8}$, a constant learning rate of $\eta = 0.001$, and a weight decay of $10e^{-6}$.  
We applied Xavier initialization to initialize the weights of all models and minibatch training with a batch size of 128, sampled randomly for 60,000 iterations in all PDE cases.

In the MLP models, we tested two variants with different activation functions: standard Tanh and parametric-Tanh. The standard model used Tanh as its activation function, while the latter used Tanh$(ax+b)$, where $a$ and $b$ are learnable parameters initialized to $1$ and $0$, respectively.
We evaluate the performance of the models using the relative $L_2$-norm (Eq. ~\ref{eq:l2_norm}) and measure the time per iteration in seconds. 

\begin{equation}
    \text{Relative } L_2 \text{ norm} = \frac{\| y - \hat{y} \|_2}{\| y \|_2} * 100\%
    \label{eq:l2_norm}
\end{equation}

\noindent For simplicity, throughout the remainder of the paper, we refer to the models by their activation or basis function along with the respective architecture identifier. For example, Tanh(A1) represents an MLP with Tanh activation function using A1 architecture, whereas B-spline(A2) represents a KAN model with B-spline basis function using architecture A2.
Although our results could be further optimized through more extensive hyperparameter tuning or an exhaustive architecture search, our focus in this work is on evaluating the models under consistent settings.

Finally, all experiments are performed on an NVIDIA A100 system consisting of a single node with 4 GPUs, each with 40 GB of VRAM.


\subsection{Results}
\label{sec:results}

\begin{figure}[t]
    \centering
    \includegraphics[width=0.85\columnwidth, trim={0cm 0cm 0cm 0cm}, clip]{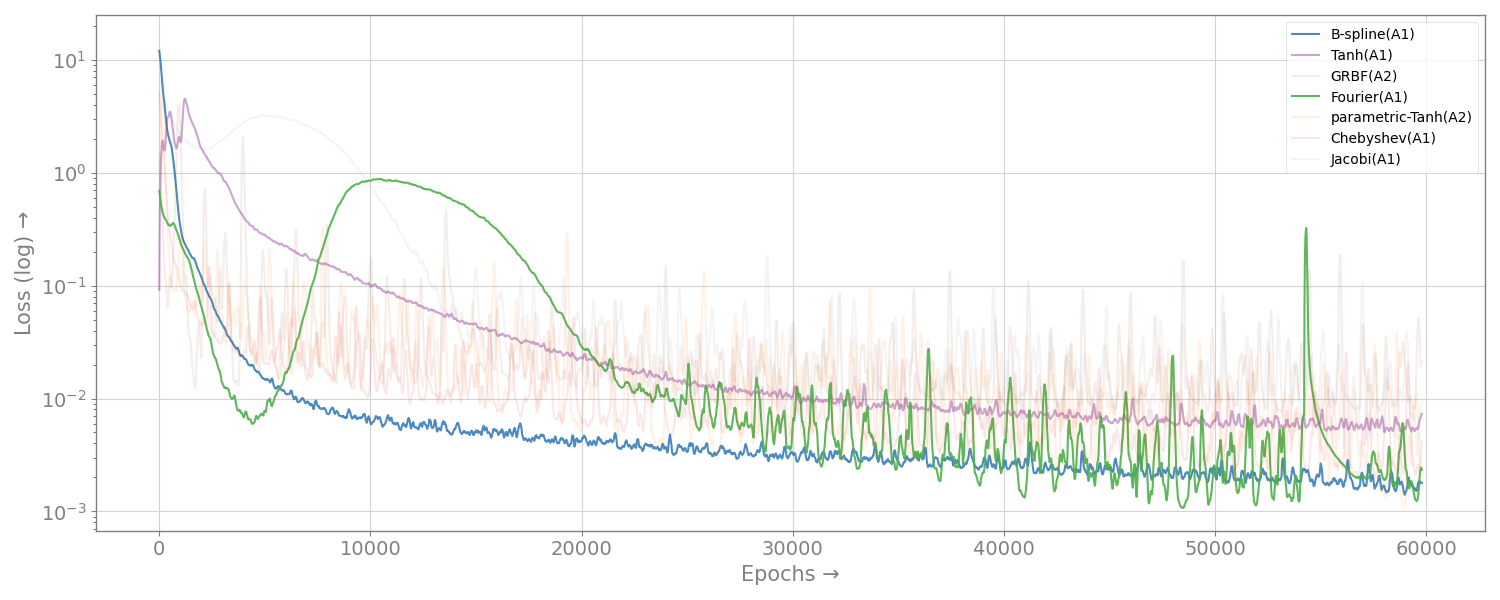} 
    \caption{Training convergence history for the Helmholtz equation. The loss curves for all models are shown, with \bspline{B-spline(A1)}, \fourier{Fourier(A1)}, and \tanhc{Tanh(A1)} highlighted in saturated colors, which demonstrated the best convergence behavior. The remaining models are depicted with transparency to enhance visual distinction and clarity.
    }
    \label{fig:loss_history_helmholtz}
\end{figure}

\begin{figure}[t]
    \centering
    \begin{tabular} {
        c@{\hspace{1.0pt}} @{\hspace{1.0pt}}
        c@{\hspace{1.0pt}} @{\hspace{1.0pt}}
        c@{\hspace{1.0pt}} @{\hspace{1.0pt}}
        c@{\hspace{1.0pt}} @{\hspace{1.0pt}}
        c@{\hspace{1.0pt}} @{\hspace{1.0pt}}
    }
        
        \includegraphics[width=0.195\columnwidth, trim={0.35cm 0.35cm 0.25cm 0.25cm}, clip]{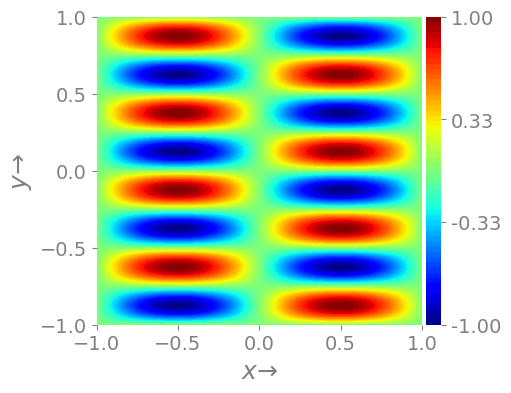} &
        \includegraphics[width=0.195\columnwidth,trim={0.35cm 0.35cm 0.25cm 0.25cm}, clip]{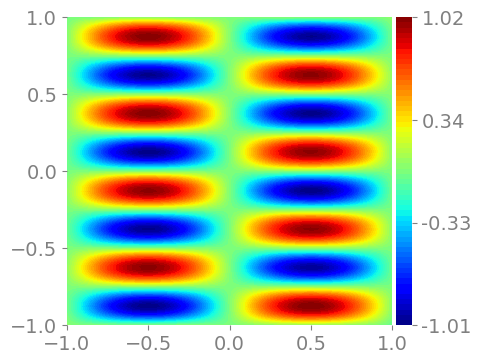} &
        \includegraphics[width=0.195\columnwidth,trim={0.35cm 0.35cm 0.25cm 0.25cm}, clip]{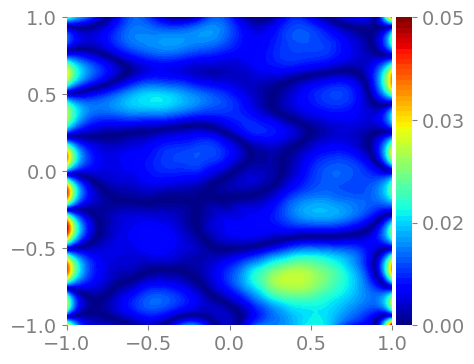} &
        \includegraphics[width=0.195\columnwidth,trim={0.35cm 0.35cm 0.25cm 0.25cm}, clip]{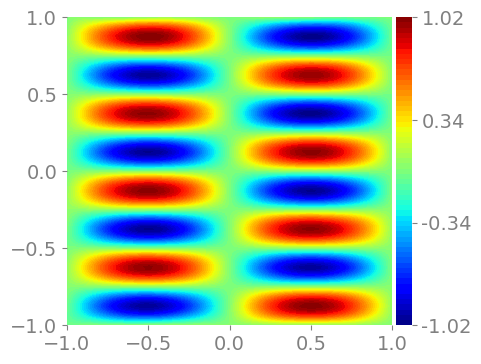} &
        \includegraphics[width=0.195\columnwidth,trim={0.35cm 0.35cm 0.25cm 0.25cm}, clip]{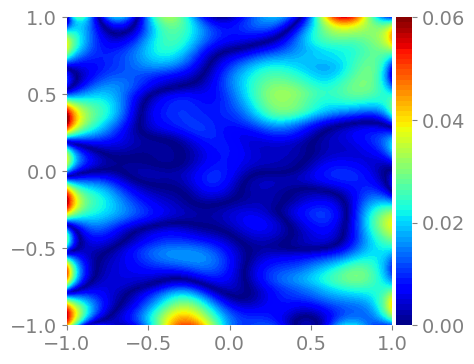} \\
        
        \tiny (a) reference & \tiny (b) prediction of B-spline(A1) & \tiny (c) error $|(b) - (a)|$  & \tiny (d) prediction of Fourier(A1) & \tiny (e) error $|(d) - (a)|$\\
       
    \end{tabular}
    \caption{Contour plots for the solution of the Helmholtz equation. 
    The plots depict the reference or actual solution, the predicted solution, and the corresponding error for the two best-performing models identified in Table~\ref{table:results}. 
    (b) presents the prediction of \bspline{B-spline(A1)} model, which achieves the lowest error of 1.93\%, while (d) shows the prediction of \fourier{Fourier(A1)} model, which has a relative error of 3.09\%.
    }
    \label{fig:veloctiy_helmholtz}
\end{figure}

\begin{figure}[t]
    \centering
    \includegraphics[width=0.85\columnwidth, trim={0cm 0cm 0cm 0cm}, clip]{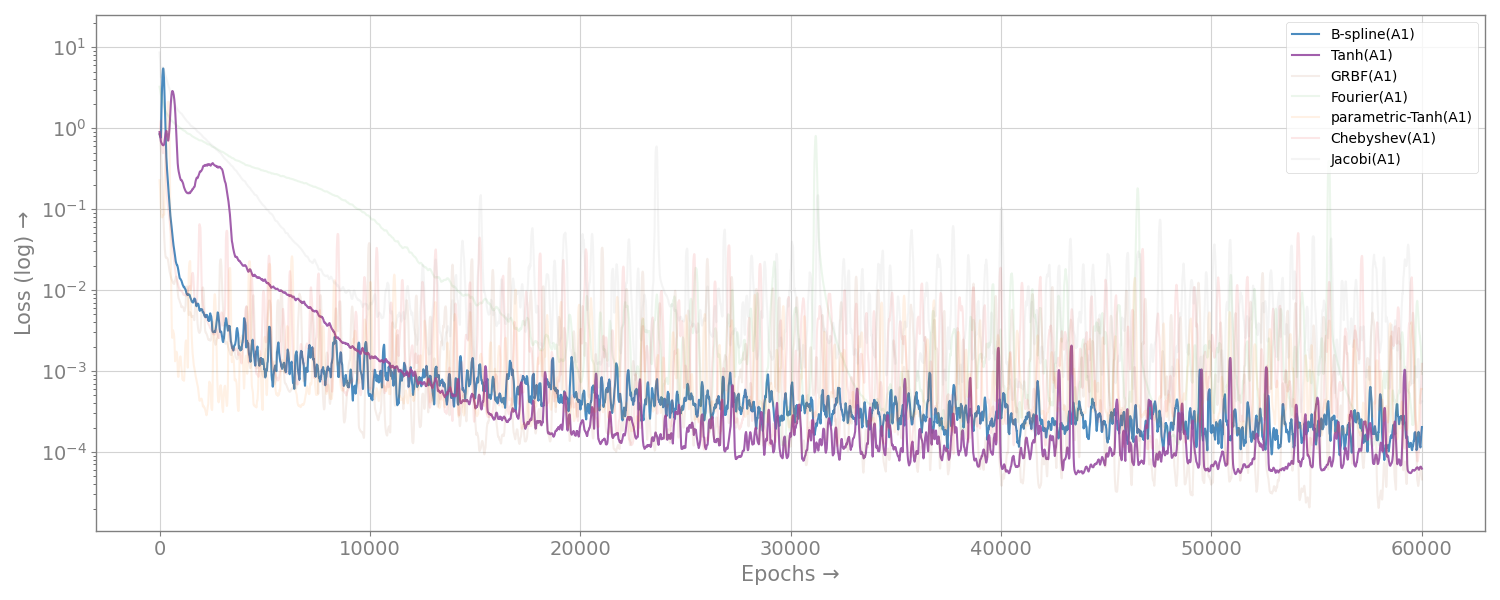} 
    \caption{Training convergence history for the Klein-Gordon equation. The loss curves for the two best converging models, \tanhc{Tanh(A1)} and \bspline{B-spline(A1)}, are highlighted.
    }
    \label{fig:loss_history_Klein_Gordon}
\end{figure}

\begin{figure}[t]
    \centering
    \begin{tabular}{
        c@{\hspace{1.5pt}} @{\hspace{1.5pt}}
        c@{\hspace{1.5pt}} @{\hspace{1.5pt}}
        c@{\hspace{1.5pt}} @{\hspace{1.5pt}}
        c@{\hspace{1.5pt}} @{\hspace{1.5pt}}
        c@{\hspace{1.5pt}} @{\hspace{1.5pt}}
    }
        \includegraphics[width=0.195\columnwidth, trim={0.20cm 0.20cm 0.20cm 0.20cm}, clip]{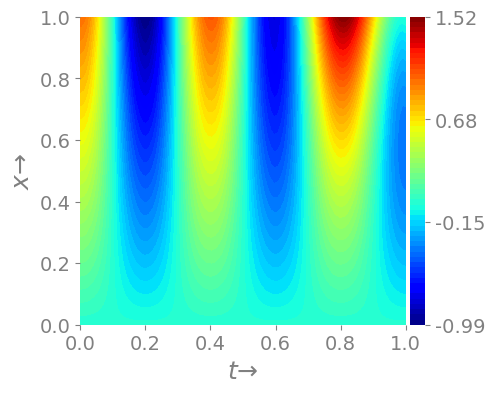} &
        \includegraphics[width=0.195\columnwidth, trim={0.20cm 0.20cm 0.20cm 0.20cm}, clip]{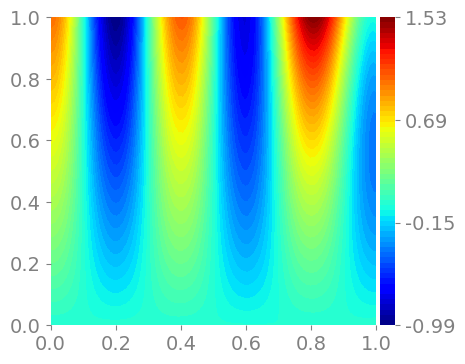} &
        \includegraphics[width=0.195\columnwidth, trim={0.20cm 0.20cm 0.20cm 0.20cm}, clip]{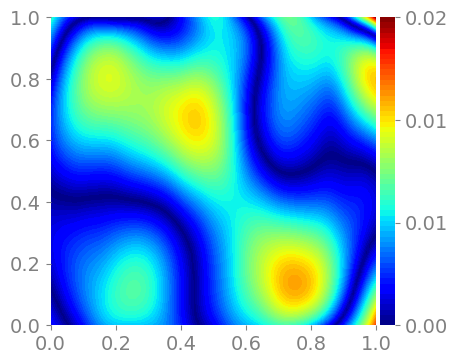} &
        \includegraphics[width=0.195\columnwidth, trim={0.20cm 0.20cm 0.20cm 0.20cm}, clip]{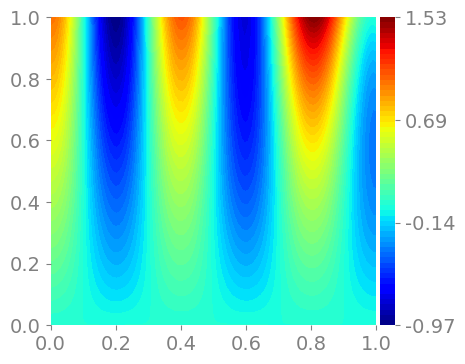} &
        \includegraphics[width=0.195\columnwidth, trim={0.20cm 0.20cm 0.20cm 0.20cm}, clip]{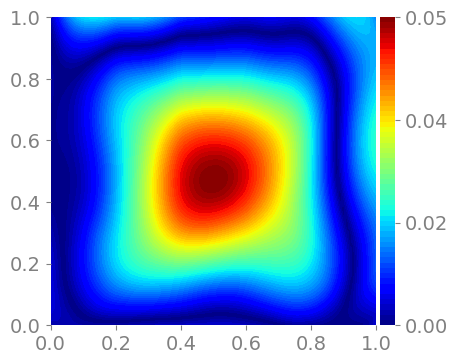} \\
        
        \tiny (a) reference & \tiny (b) prediction of Tanh(A1) & \tiny (c) error $|(b) - (a)|$ & \tiny (d) prediction of B-spline(A1) & \tiny (e) error $|(d) - (a)|$ \\
        
        \end{tabular}
        \caption{Contour plots for the solution of the Klein-Gordon equation.
        The plots depict the reference or actual solution, the predicted solution, and the corresponding error for the two best-performing models identified in Table~\ref{table:results}. 
        (b) presents the prediction of \tanhc{Tanh(A1)} model, which achieves the lowest error of 1.58\%, while (d) shows the prediction of \bspline{B-spline(A1)} model, which has a relative error of 5.23\%.
        }
        
        \label{fig:veloctiy_Klein}
\end{figure}

\begin{figure}[t]
    \centering
    \includegraphics[width=0.85\columnwidth, trim={0cm 0cm 0cm 0cm}, clip]{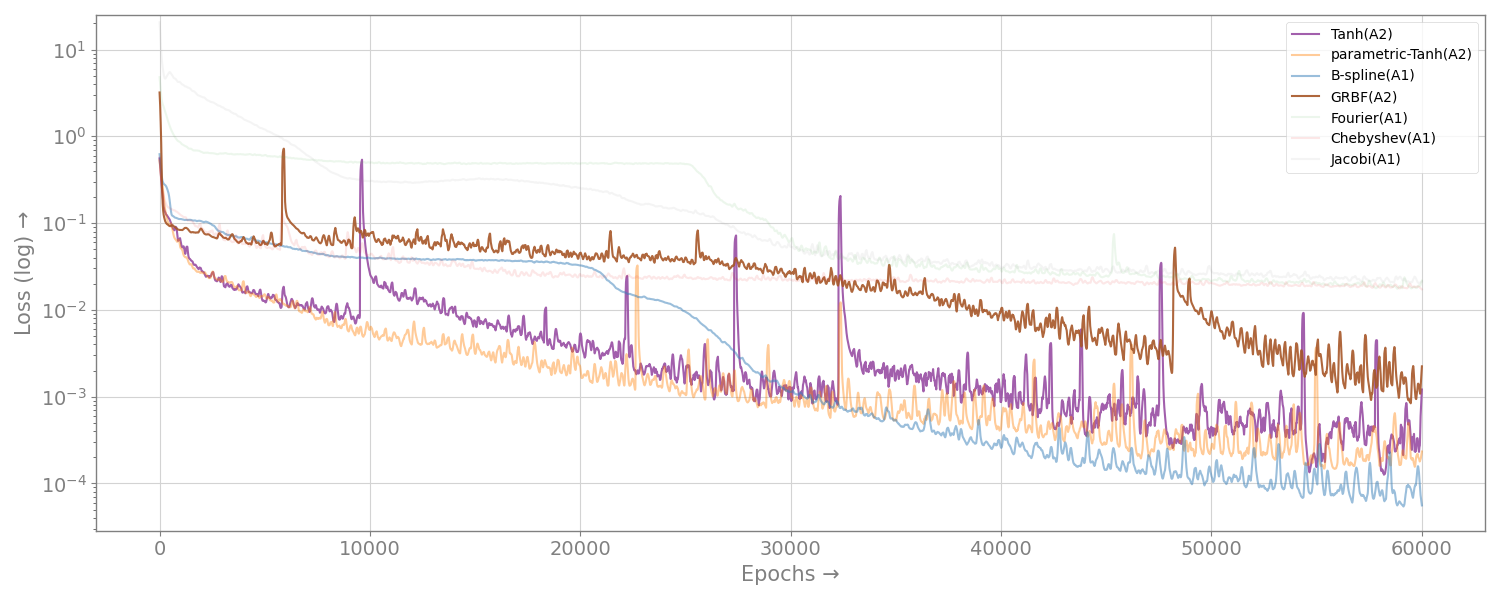} 
    \caption{Training convergence history for the Wave equation. The loss curves indicate that \bspline{B-spline(A1)} and \ptanhc{parametric-Tanh(A2)} models exhibit the most consistent convergence. However, \tanhc{Tanh(A2)} and \grbf{GRBF(A2)} models achieve the lowest testing errors, as reported in Table~\ref{table:results}.
    }
    \label{fig:loss_history_wave}
\end{figure}

\begin{figure}[t]
    \centering
    \begin{tabular}{
        c@{\hspace{1.5pt}} @{\hspace{1.5pt}}
        c@{\hspace{1.5pt}} @{\hspace{1.5pt}}
        c@{\hspace{1.5pt}} @{\hspace{1.5pt}}
        c@{\hspace{1.5pt}} @{\hspace{1.5pt}}
        c@{\hspace{1.5pt}} @{\hspace{1.5pt}}
    }
        \includegraphics[width=0.195\columnwidth, trim={0.20cm 0.20cm 0.20cm 0.20cm}, clip]{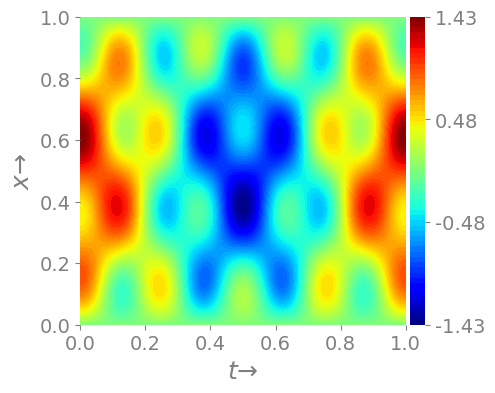} &
        \includegraphics[width=0.195\columnwidth, trim={0.20cm 0.20cm 0.20cm 0.20cm}, clip]{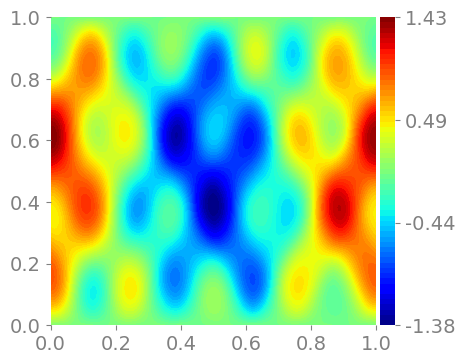} &
        \includegraphics[width=0.195\columnwidth, trim={0.20cm 0.20cm 0.20cm 0.20cm}, clip]{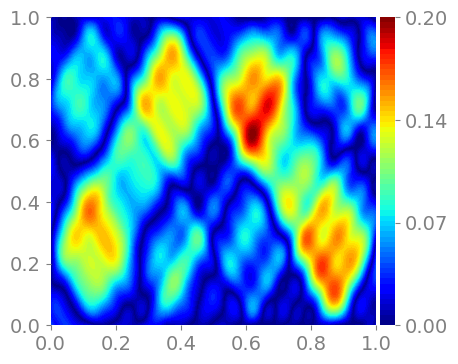} &
        \includegraphics[width=0.195\columnwidth, trim={0.20cm 0.20cm 0.20cm 0.20cm}, clip]{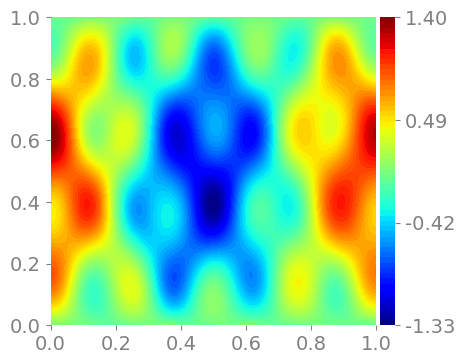} &
        \includegraphics[width=0.195\columnwidth, trim={0.20cm 0.20cm 0.20cm 0.20cm}, clip]{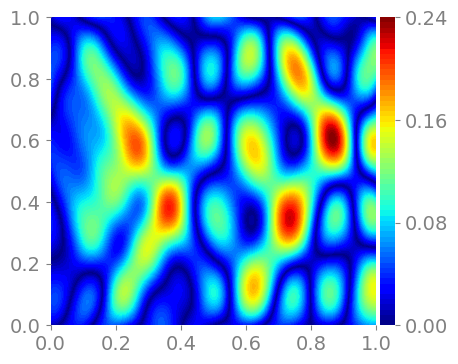} \\
        
        \tiny (a) reference & \tiny (b) prediction of Tanh(A2) & \tiny (c) error $|(b) - (a)|$  & \tiny (d) prediction of GRBF(A2) & \tiny (e) error $|(d) - (a)|$ \\
        
    \end{tabular}
    \caption{Contour plots for the solution of the Wave equation in the spatio-temporal domain $x$. The plots depict the reference or actual solution, the predicted solution, and the corresponding error for the two best-performing models identified in Table~\ref{table:results}. (b) presents the prediction of \tanhc{Tanh(A2)} model, which achieves the lowest error of 13.6\%, while (d) shows the prediction of \grbf{GRBF(A2)} model, which has a relative error of 14.7\%.
    }
        \label{fig:veloctiy_wave}
\end{figure}

\begin{figure}[t]
    \centering
    \includegraphics[width=0.85\columnwidth, trim={0cm 0cm 0cm 0cm}, clip]{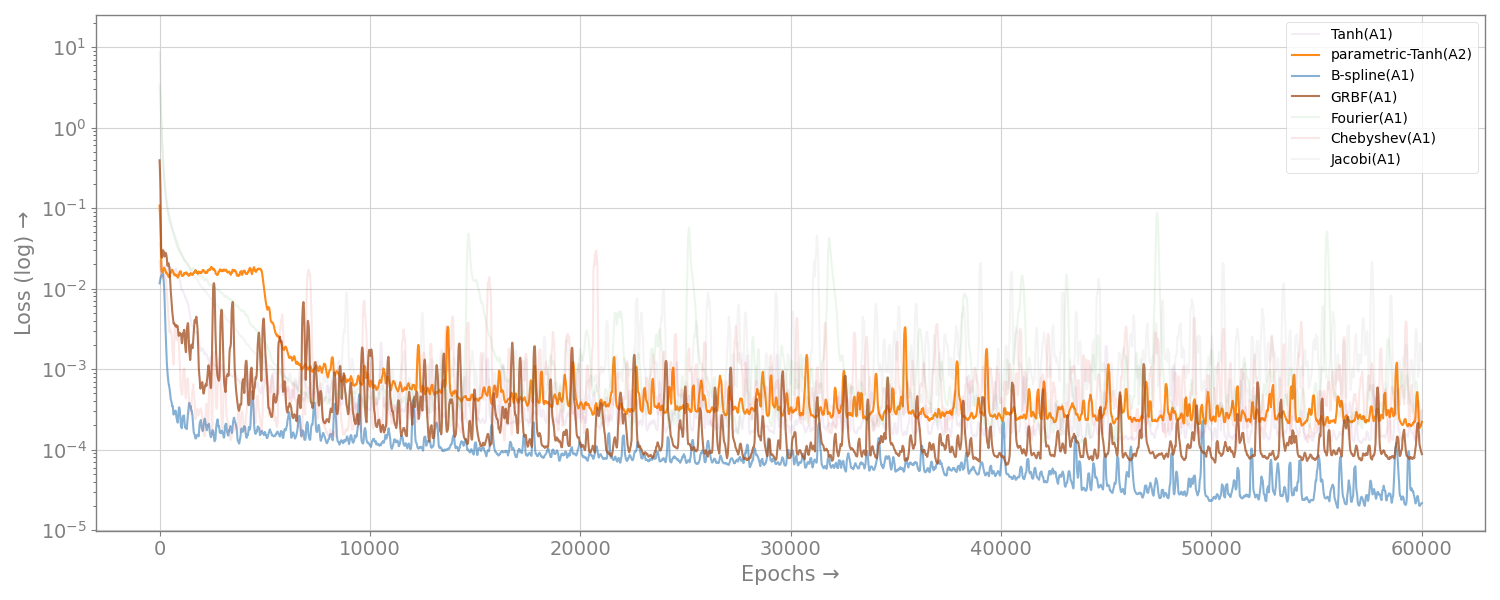} 
    \caption{Training convergence history for the Convection-diffusion equation.
    The loss curves for all models are displayed, with \bspline{B-spline(A1)}, \grbf{GRBF(A1)}, and \ptanhc{parametric-Tanh(A2)} demonstrated the best convergence. 
    }
    
    \label{fig:loss_history_Diffusion}
\end{figure}

\begin{figure}[t]
    \centering
    \begin{tabular}{
    >{\centering\arraybackslash}m{1.5cm} 
    >{\centering\arraybackslash}m{3cm} 
    >{\centering\arraybackslash}m{3cm} 
    >{\centering\arraybackslash}m{3cm} 
    >{\centering\arraybackslash}m{3cm} 
    }
        &\tiny{$u$ at $t_0$} & \tiny{$u$ at $t_{1}$} & \tiny{$f$ at $t_0$} & \tiny{$f$ at $t_{1}$} \\
         
        \tiny{\makecell{(a) reference}} &
        \includegraphics[width=0.195\columnwidth, trim={0.20cm 0.20cm 0.20cm 0.20cm}, clip]{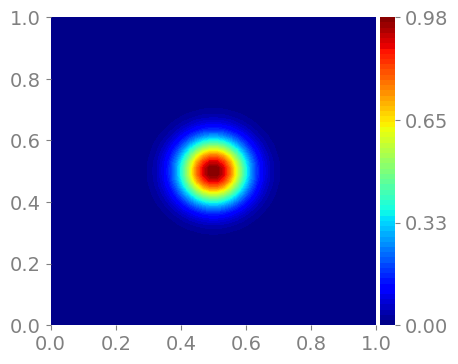}  &
        \includegraphics[width=0.195\columnwidth, trim={0.20cm 0.20cm 0.20cm 0.20cm}, clip]{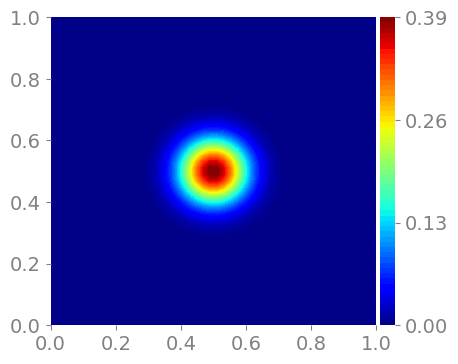}  &
        \includegraphics[width=0.195\columnwidth, trim={0.20cm 0.20cm 0.20cm 0.20cm}, clip]{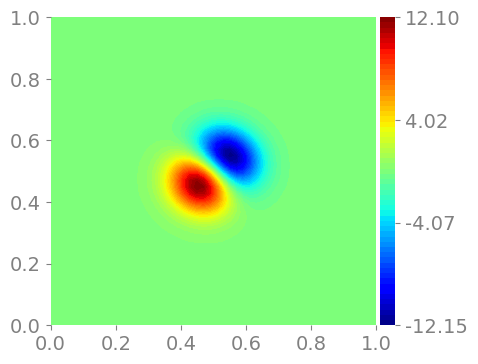}  &
        \includegraphics[width=0.195\columnwidth, trim={0.20cm 0.20cm 0.20cm 0.20cm}, clip]{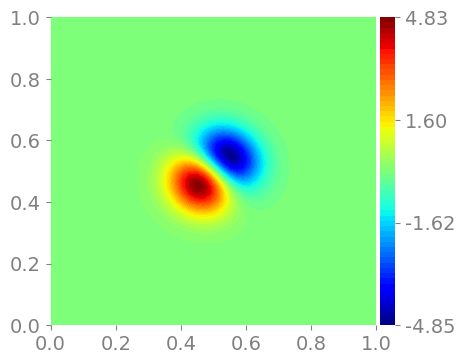} \\
        
        \tiny{\makecell{(b) prediction of \\ GRBF(A1)}}&\includegraphics[width=0.195\columnwidth, trim={0.35cm 0.35cm 0.20cm 0.20cm}, clip]{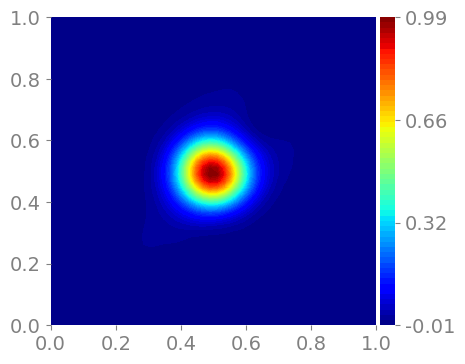}  &
        \includegraphics[width=0.195\columnwidth, trim={0.20cm 0.20cm 0.20cm 0.20cm}, clip]{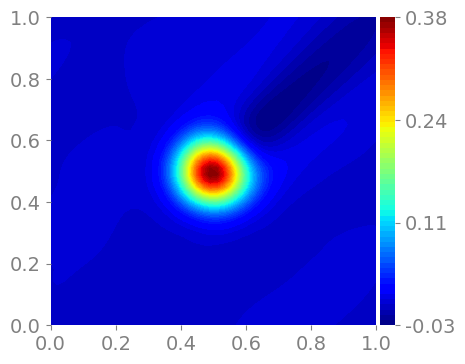}  &
        \includegraphics[width=0.195\columnwidth, trim={0.20cm 0.20cm 0.20cm 0.20cm}, clip]{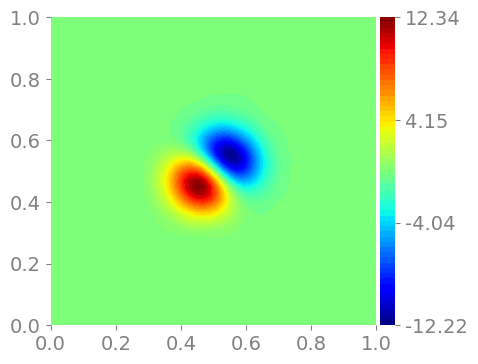}  &
        \includegraphics[width=0.195\columnwidth, trim={0.20cm 0.20cm 0.20cm 0.20cm}, clip]{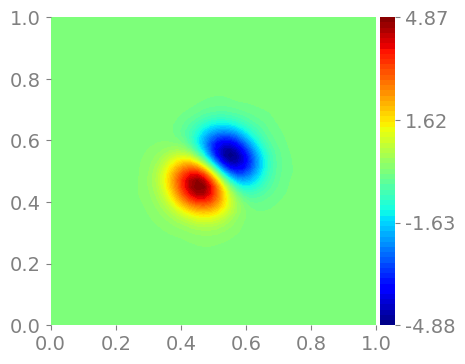} \\
        
        \tiny{\makecell{ (c) error $|(b) - (a)|$}} &
        \includegraphics[width=0.195\columnwidth, trim={0.20cm 0.20cm 0.20cm 0.20cm}, clip]{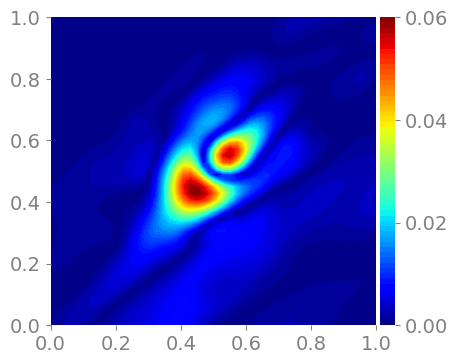}  &
        \includegraphics[width=0.195\columnwidth, trim={0.20cm 0.20cm 0.20cm 0.20cm}, clip]{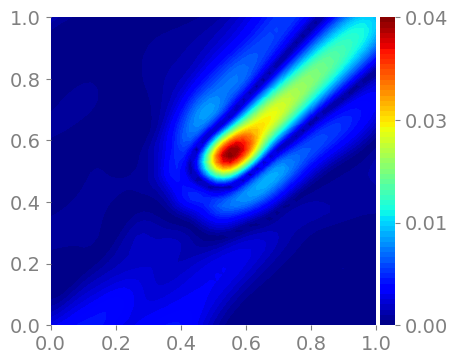}  &
        \includegraphics[width=0.195\columnwidth, trim={0.20cm 0.20cm 0.20cm 0.20cm}, clip]{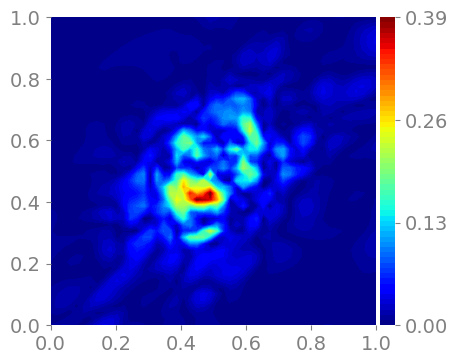}  &
        \includegraphics[width=0.195\columnwidth, trim={0.20cm 0.20cm 0.20cm 0.20cm}, clip]{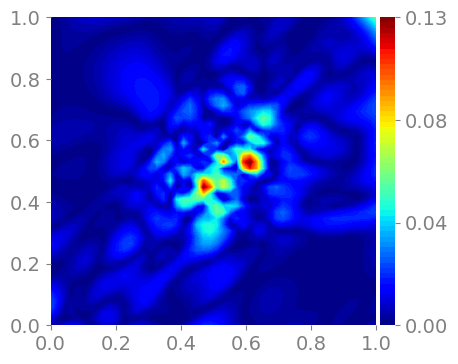} \\
        
        \tiny{\makecell{ (d) prediction of \\ param-Tanh(A2)}} & \includegraphics[width=0.195\columnwidth, trim={0.20cm 0.20cm 0.20cm 0.20cm}, clip]{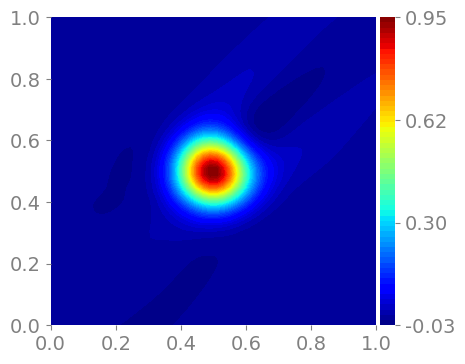}  &
        \includegraphics[width=0.195\columnwidth, trim={0.20cm 0.20cm 0.20cm 0.20cm}, clip]{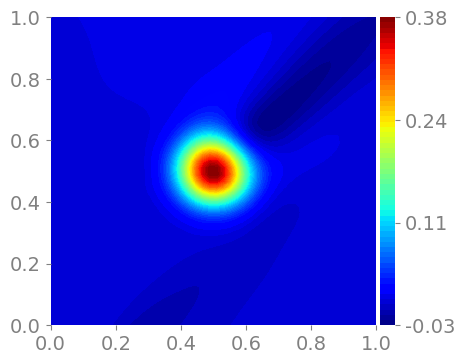}  &
        \includegraphics[width=0.195\columnwidth, trim={0.20cm 0.20cm 0.20cm 0.20cm}, clip]{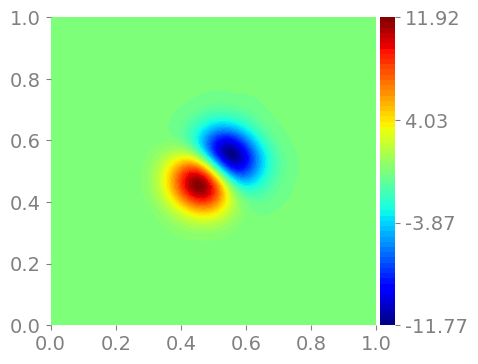}  &
        \includegraphics[width=0.195\columnwidth, trim={0.20cm 0.20cm 0.20cm 0.20cm}, clip]{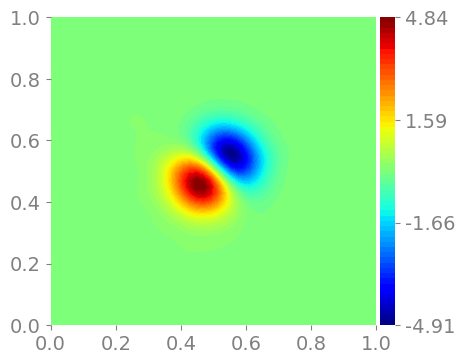} \\
        
        \tiny{\makecell{(e) error $|(d) - (a)|$}} & \includegraphics[width=0.195\columnwidth, trim={0.20cm 0.20cm 0.20cm 0.20cm}, clip]{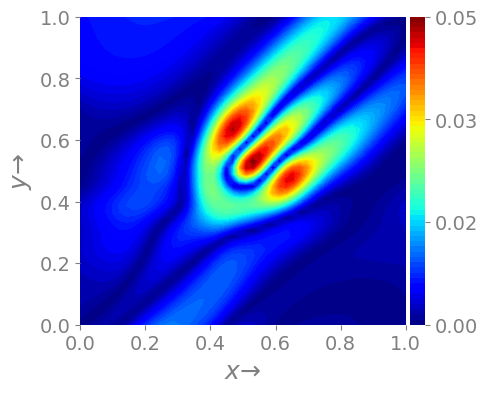}  &
        \includegraphics[width=0.195\columnwidth, trim={0.20cm 0.20cm 0.20cm 0.20cm}, clip]{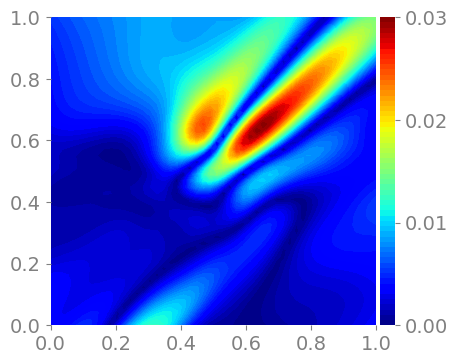}  &
        \includegraphics[width=0.195\columnwidth, trim={0.20cm 0.20cm 0.20cm 0.20cm}, clip]{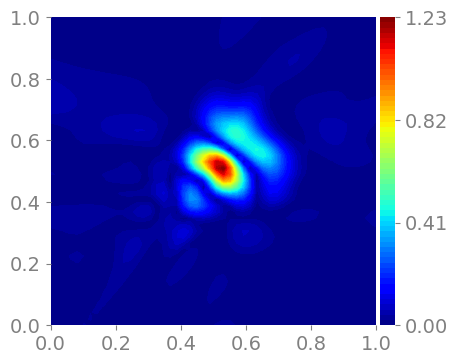}  &
        \includegraphics[width=0.195\columnwidth, trim={0.20cm 0.20cm 0.20cm 0.20cm}, clip]{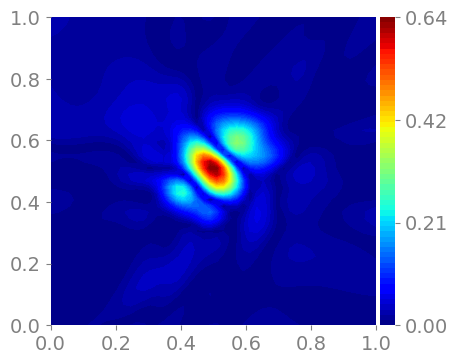} \\
        
    \end{tabular}
    \caption{
    Contour plots for the solution of the Convection-diffusion velocity ($v$) and force ($f$) values at time-step zero ($t_0$) and one ($t_1$) in the spatio-temporal domain. The plots at each row depict the reference or actual solution, the predicted solution, and the corresponding error for the two best-performing models identified in Table~\ref{table:results}, respectively. (b) presents the prediction of \grbf{GRBF(A1)} model, the relative $L_2$ errors  are $11.5\%$ $2.07\%$ for $u$ and $f$, respectively. (d) presents the prediction of \ptanhc{parametric-Tanh(A2)} model, the relative $L_2$ errors  are $11.6\%$ $1.07\%$ for $u$ and $f$, respectively.
    } 
    \label{fig:diffusion}
\end{figure}

\begin{figure}[t]
    \centering
    \includegraphics[width=0.85\columnwidth, trim={0cm 0cm 0cm 0cm}, clip]{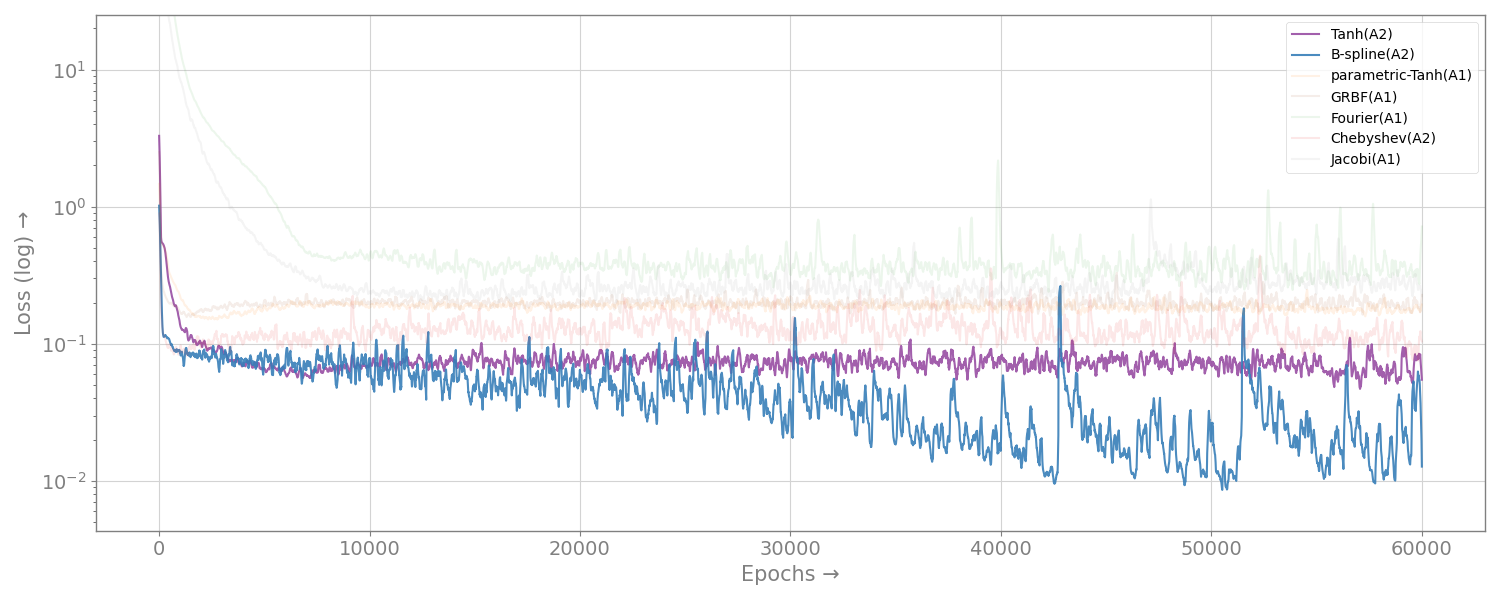} 
    \caption{Training convergence history for the Cavity problem. Loss curves for all models are shown, with \bspline{B-spline(A2)} and \tanhc{Tanh(A2)}, which demonstrated the best convergence.
   }
   \label{fig:loss_history_cavity}
\end{figure}

 \begin{figure}[t]
    \centering
    \begin{tabular}{
    >{\centering\arraybackslash}m{1cm} 
    >{\centering\arraybackslash}m{4cm} 
    >{\centering\arraybackslash}m{4cm} 
    >{\centering\arraybackslash}m{4cm}
    }
         &  \tiny{$u_x$ at $t_{0}$} & \tiny{$v_y$ at $t_{0}$} & \tiny{$p$ at $t_{0}$}  \\
       \tiny{\makecell{(a) reference}} & \includegraphics[width=0.195\columnwidth, trim={0.1cm 0.35cm 0.25cm 0.25cm}, clip]{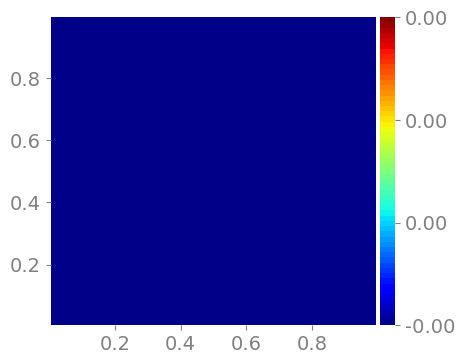}  &
        \includegraphics[width=0.195\columnwidth, trim={0.1cm 0.35cm 0.25cm 0.25cm}, clip]{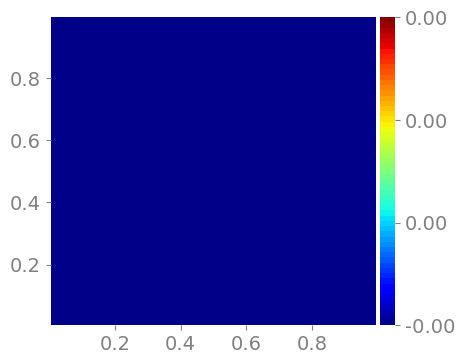}  &
        \includegraphics[width=0.195\columnwidth, trim={0.1cm 0.35cm 0.25cm 0.25cm}, clip]{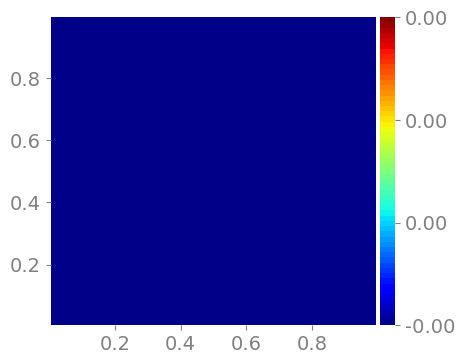}  \\
        \tiny{\makecell{(b) prediction of \\ Tanh(A2)}} & \includegraphics[width=0.195\columnwidth, trim={0.1cm 0.35cm 0.25cm 0.25cm}, clip]{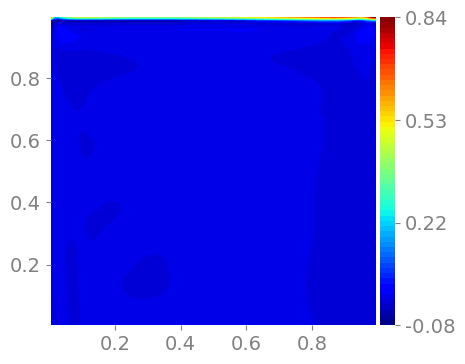}  &
        \includegraphics[width=0.195\columnwidth, trim={0.1cm 0.35cm 0.25cm 0.25cm}, clip]{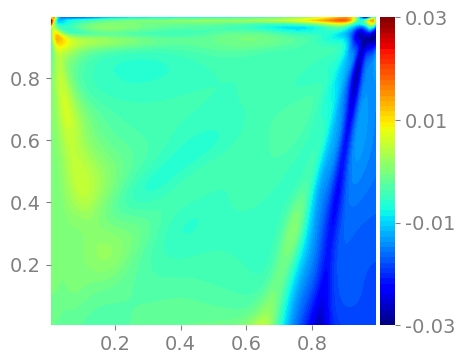}  &
        \includegraphics[width=0.195\columnwidth, trim={0.1cm 0.35cm 0.25cm 0.25cm}, clip]{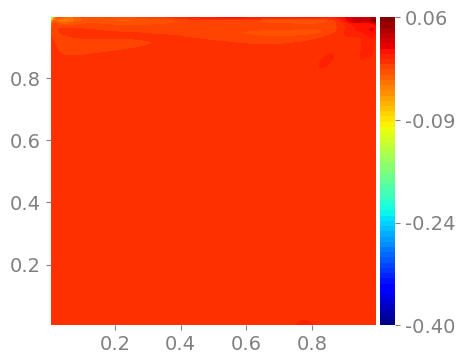}   \\
        \tiny{\makecell{(c) error $|(b) - (a)|$ }} & \includegraphics[width=0.195\columnwidth, trim={0.1cm 0.35cm 0.25cm 0.25cm}, clip]{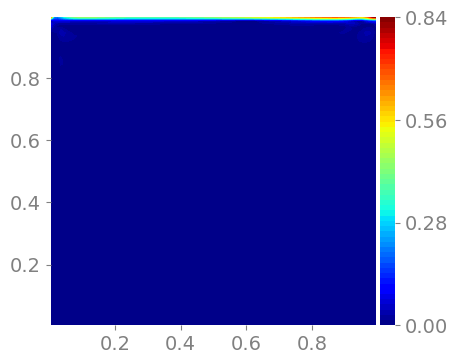}  &
        \includegraphics[width=0.195\columnwidth, trim={0.1cm 0.35cm 0.25cm 0.25cm}, clip]{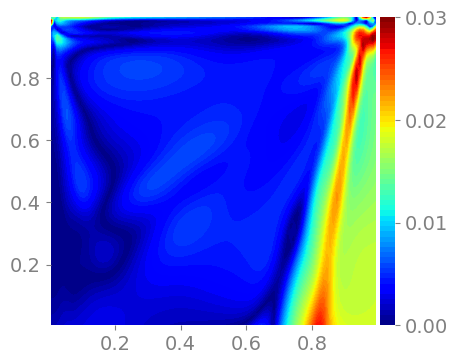}  &
        \includegraphics[width=0.195\columnwidth, trim={0.1cm 0.35cm 0.25cm 0.25cm}, clip]{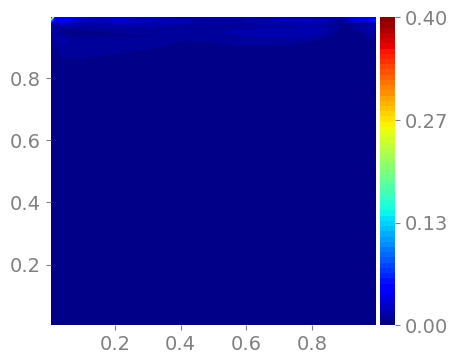}  \\
        \tiny{\makecell{(d) prediction of \\ B-spline(A2)}} & \includegraphics[width=0.195\columnwidth, trim={0.1cm 0.35cm 0.25cm 0.25cm}, clip]{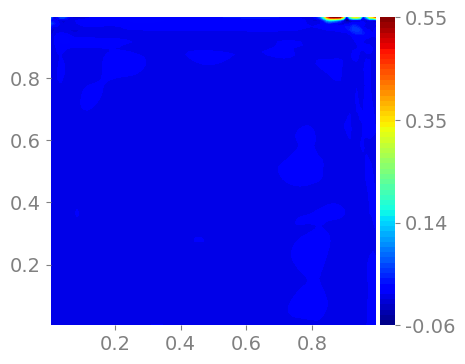}  &
        \includegraphics[width=0.195\columnwidth, trim={0.1cm 0.35cm 0.25cm 0.25cm}, clip]{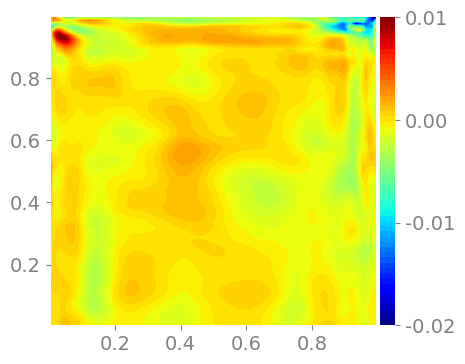}  &
        \includegraphics[width=0.195\columnwidth, trim={0.1cm 0.35cm 0.25cm 0.25cm}, clip]{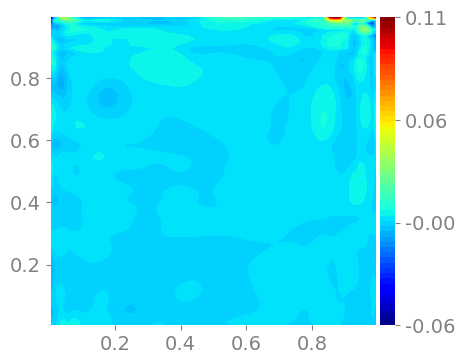} \\
        \tiny{\makecell{(e) error $|(d) - (a)|$}} & \includegraphics[width=0.195\columnwidth, trim={0.1cm 0.35cm 0.25cm 0.25cm}, clip]{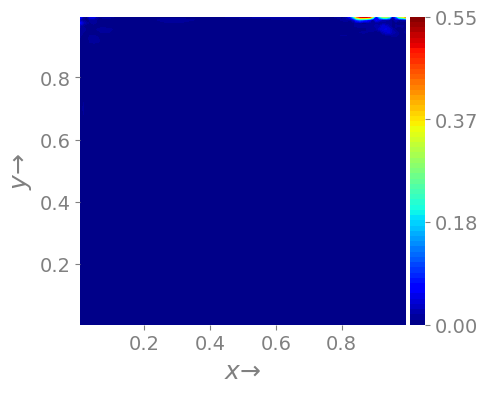}  &
        \includegraphics[width=0.195\columnwidth, trim={0.1cm 0.35cm 0.25cm 0.25cm}, clip]{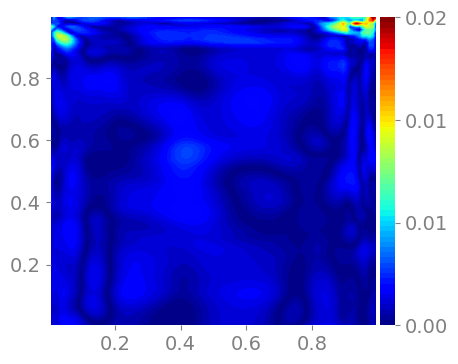}  &
        \includegraphics[width=0.195\columnwidth, trim={0.1cm 0.35cm 0.25cm 0.25cm}, clip]{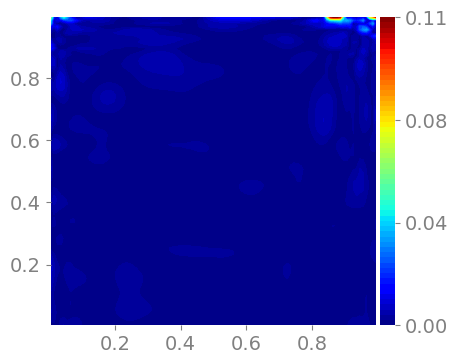} \\
    \end{tabular}
    \caption{ 
    Contour plots for the solution of the Cavity problem velocity ($u_x$ and $v_y$) and pressure ($p$) values at the initial condition($t_0$), with $Re = 100$. 
    (b) presents the prediction of \tanhc{Tanh(A2)} model, the relative $L_2$ errors as 10.8\%, 22\%, and 24.4\% for $u_x$, $v_y$, and $p$, respectively. (d) presents the prediction of \bspline{B-spline(A2)} model, the relative $L_2$ errors are 4.1\%, 8.7\%, and 21.3\% for $u_x$, $v_y$, and $p$, respectively.
    }
    \label{fig:cavity_time_0}
\end{figure}

\begin{figure}[t]
    \centering
    \begin{tabular}{
    >{\centering\arraybackslash}m{1cm} 
    >{\centering\arraybackslash}m{4cm} 
    >{\centering\arraybackslash}m{4cm} 
    >{\centering\arraybackslash}m{4cm}
    }
         &  \tiny{$u_x$ at $t_{99}$} & \tiny{$v_x$ at $t_{99}$} & \tiny{$p$ at $t_{99}$} \\
        \tiny{\makecell{(a) reference}} & \includegraphics[width=0.195\columnwidth, trim={0.1cm 0.35cm 0.25cm 0.25cm}, clip]{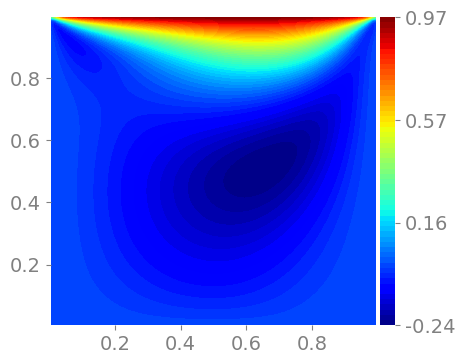}  &
        \includegraphics[width=0.195\columnwidth, trim={0.1cm 0.35cm 0.25cm 0.25cm}, clip]{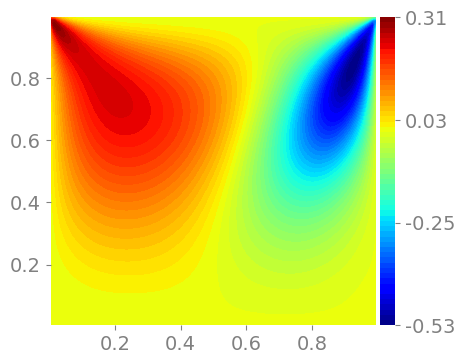}  &
        \includegraphics[width=0.195\columnwidth, trim={0.1cm 0.35cm 0.25cm 0.25cm}, clip]{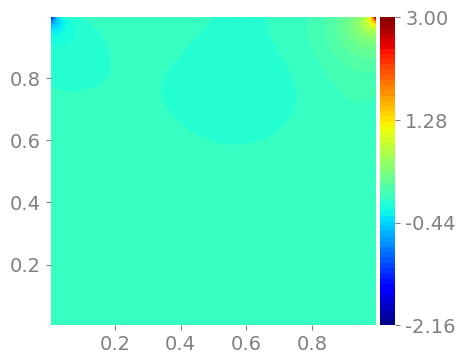}  \\
        \tiny{\makecell{(b) prediction of \\ Tanh(A2)}} & \includegraphics[width=0.195\columnwidth, trim={0.1cm 0.35cm 0.25cm 0.25cm}, clip]{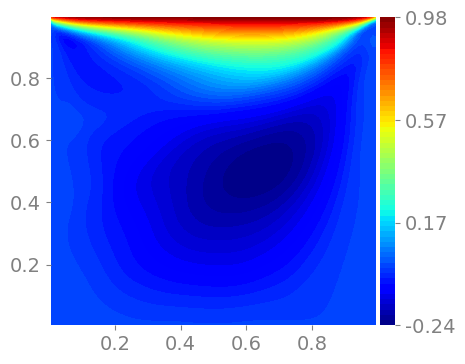}  &
        \includegraphics[width=0.195\columnwidth, trim={0.1cm 0.35cm 0.25cm 0.25cm}, clip]{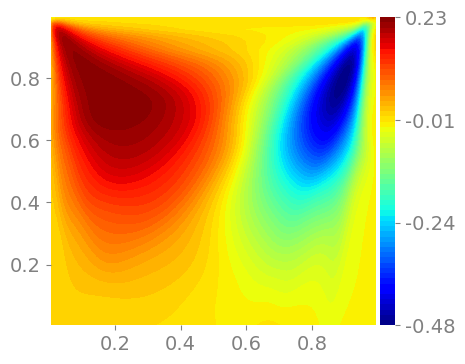}  &
        \includegraphics[width=0.195\columnwidth, trim={0.1cm 0.35cm 0.25cm 0.25cm}, clip]{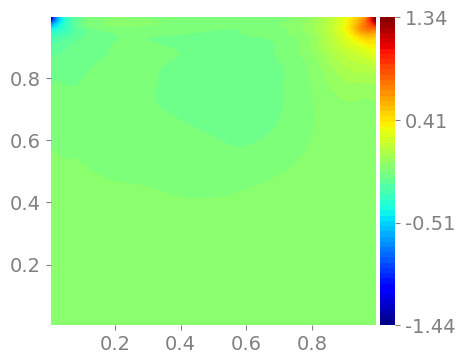}   \\
        \tiny{\makecell{(c) error $|(b) - (a)|$}}&\includegraphics[width=0.195\columnwidth, trim={0.1cm 0.35cm 0.25cm 0.25cm}, clip]{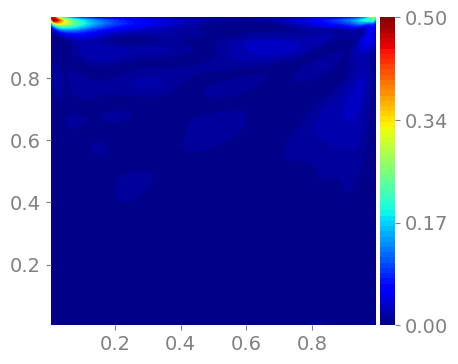}  &
        \includegraphics[width=0.195\columnwidth, trim={0.1cm 0.35cm 0.25cm 0.25cm}, clip]{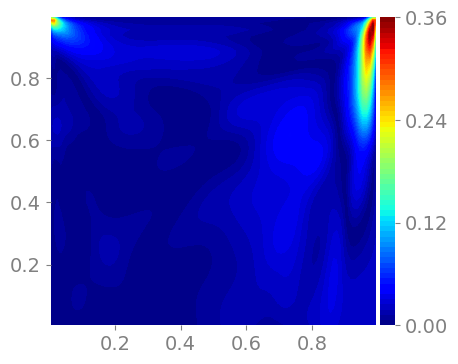}  &
        \includegraphics[width=0.195\columnwidth, trim={0.1cm 0.35cm 0.25cm 0.25cm}, clip]{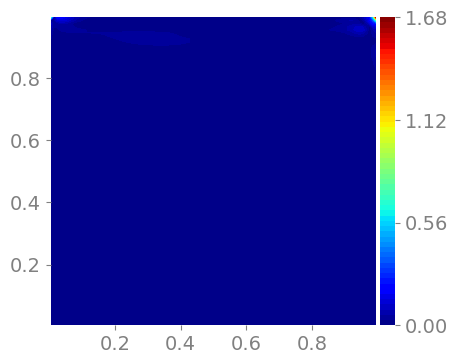}  \\
        \tiny{\makecell{(d) prediction of \\ B-spline(A2)}} & \includegraphics[width=0.195\columnwidth, trim={0.1cm 0.35cm 0.25cm 0.25cm}, clip]{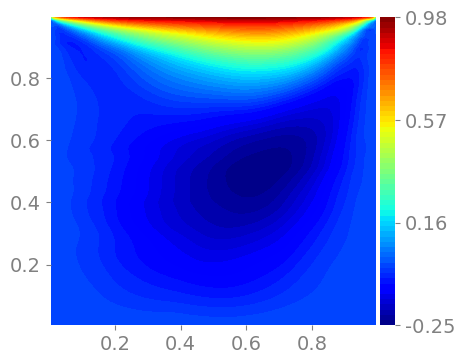}  &
        \includegraphics[width=0.195\columnwidth, trim={0.1cm 0.35cm 0.25cm 0.25cm}, clip]{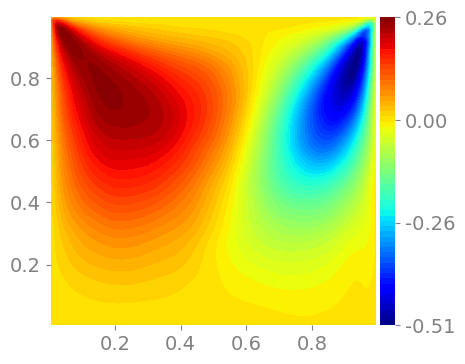}  &
        \includegraphics[width=0.195\columnwidth, trim={0.1cm 0.35cm 0.25cm 0.25cm}, clip]{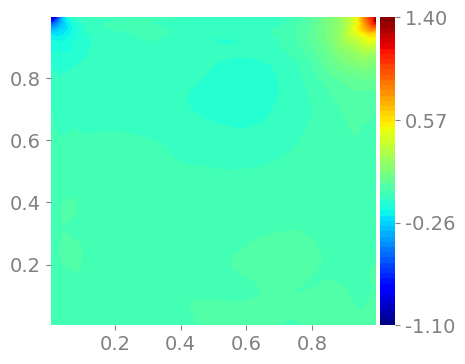} \\
        \tiny{\makecell{(e) error $|(d) - (a)|$}} & \includegraphics[width=0.195\columnwidth, trim={0.1cm 0.35cm 0.25cm 0.25cm}, clip]{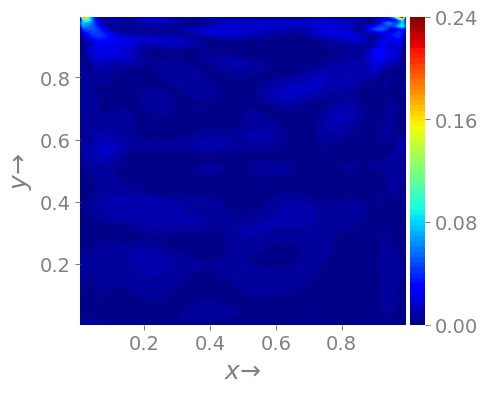}  &
        \includegraphics[width=0.195\columnwidth, trim={0.1cm 0.35cm 0.25cm 0.25cm}, clip]{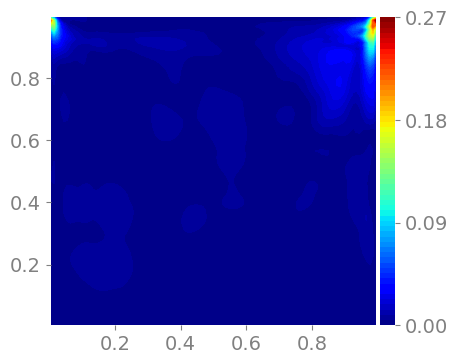}  &
        \includegraphics[width=0.195\columnwidth, trim={0.1cm 0.35cm 0.25cm 0.25cm}, clip]{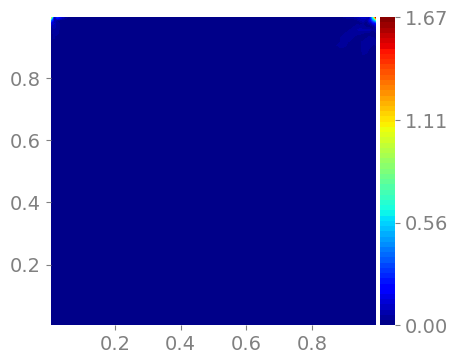} \\
        
    \end{tabular}
    \caption{ 
    Contour plots for the solution of the Cavity problem, showing velocity components ($u_x$ and $v_y$) and pressure ($p$) values at time-step 99 ($t_{99}$), with $Re = 100$. 
    (b) shows the prediction from the \tanhc{Tanh(A2)} model, with relative $L_2$ errors of 10.8\% 22\%  24.4\% for $u_x$, $v_y$, and $p$, respectively. (d) shows the prediction from the \ptanhc{B-spline(A2)} model, which achieved relative $L_2$ errors of 4.1\%, 8.7\%, 21.3\% for $u_x$, $v_y$, and $p$, respectively. Note that only the result of $t_0$ is recorded in Table~\ref{table:results}, results of $t_{99}$ can be reproduced using the pretrained model provided with our source code.
    } 
    \label{fig:cavity_time_99}
\end{figure}

Table~\ref{table:results} summarizes the performance results, measured by relative $L_2$ error, for various models and architectures across different test cases. Table~\ref{table:final-loss} shows the detailed final training losses for each loss term.

\subsubsection{Helmholtz Equation}

For the Helmholtz equation, B-spline(A1) achieves the lowest velocity error (1.93\%), followed by the Tanh(A1) (6.13\%). 
In contrast, GRBF(A1) and Jacobi(A1) result in significantly higher errors. KAN models with A2 architecture exhibit instability and failure to solve the Helmholtz equation.

Figure~\ref{fig:loss_history_helmholtz} displays the training loss history for the top-performing architectures from each model group listed in Table~\ref{table:network_architecture}. The plot indicates that the Tanh(A1), Fourier(A1), and B-spline(A1) models exhibit significantly better convergence than the others, with the B-spline(A1) model achieving the lowest total final loss value. Detailed term loss values can be found in Table~\ref{table:final-loss}. The higher training error observed in the physics loss term for this case, as well as the others, reflects the weighting of the loss values discussed in section~\ref{sec:use-cases}. Nevertheless, the physics loss term remains an essential component, functioning as a regularizer.


Fig.~\ref{fig:veloctiy_helmholtz} compares the reference solution and the predicted solutions from B-spline(A1) and Fourier(A1) for the Helmholtz equation.  
The majority of inference errors for both models are concentrated near the boundaries of the domain, with the Fourier model showing more pronounced deviations in these regions.

\subsubsection{Klein-Gordon Equation}
The Tanh(A1) model has the lowest velocity error (1.58\%), followed by the B-spline(A1) model (5.23\%). Similar to the Helmholtz case, KAN models utilizing the A2 architecture show instability, with errors surpassing 90.0\% in Fourier, Chebyshev, and Jacobi, indicating a failure in performance.

Fig.~\ref{fig:loss_history_Klein_Gordon} illustrates the training convergence history for the Klein-Gordon equation. The Tanh(A1) model attained the lowest relative error, while the B-spline(A1) exhibited a more stable convergence with minimal fluctuations. The models stabilize with total training errors of approximately 0.19 for Tanh(A1) and 0.30 for B-spline(A1), as shown in Table~\ref{table:final-loss}. In contrast, other models exhibited instability and oscillatory behavior, indicating difficulties in optimizing the solution under the specified settings.

Fig.~\ref{fig:veloctiy_Klein} presents the solution of the Klein-Gordon equation, comparing the reference solution to the best-predicted solutions from Tanh(A1) and B-spline(A1), along with their corresponding error plots. Both models exhibit error concentration in regions with high gradients, where the solution changes rapidly. 
However, the Tanh(A1) handles these high-gradient areas more effectively, as indicated by the lower overall error in Fig.~\ref{fig:veloctiy_Klein}(c) compared to the error observed for the B-spline results in Fig.~\ref{fig:veloctiy_Klein}(e).

\subsubsection{Wave Equation} 
The Tanh(A2) model achieves the lowest velocity error (13.6\%), followed by GRBF(A2) (14.7\%) (as in Table~\ref{table:results}).  
Fig.~\ref{fig:loss_history_wave} presents the training convergence history for the Wave equation.
The plots indicate that B-spline(A1), Tanh(A2), and GRBF(A2) demonstrate better convergence.
The B-spline(A1) model experienced a slower convergence rate during the initial half of the training period, after which it began to converge more rapidly, ultimately reaching the lowest loss value.
The parametric-Tanh(A2) showed steady convergence, while both Tanh(A2) and GRBF(A2) exhibited some fluctuations throughout the training process.


Fig.~\ref{fig:veloctiy_wave} presents the contour plot predictions for the velocity fields using the Tanh(A2) and GRBF(A1) models. The Tanh(A2) model effectively captures the continuous dynamics of the wave equation. In contrast, the GRBF(A1) model exhibits a slightly higher relative error compared to Tanh(A2), with most of the errors occurring in areas where the wave oscillates rapidly.

\subsubsection{Convection-diffusion Equation}
The GRBF(A1) model achieves the lowest velocity $u$ followed by parametric-Tanh(A2) (refer to Table~\ref{table:results}). 
In contrast, Fourier, Chebyshev, and Jacobi models exhibited high errors with the A1 architecture and failed to converge effectively with A2. 

Fig.~\ref{fig:loss_history_Diffusion} presents the training convergence history for the Convection-diffusion equation.
The plots indicate that B-spline(A1), GRBF(A1), and parametric-Tanh(A2) exhibit better convergence compared to other models. 
The B-spline (A1) model shows the most consistent convergence and the lowest training loss across both architectures, as illustrated in Table~\ref{table:final-loss}. 
In contrast, the GRBF(A1) model shows less stability and more noticeable fluctuations. 
The remaining models display considerable instability, evidenced by pronounced oscillations in their loss curves.
Although B-spline(A1) achieves the fastest convergence and lowest final loss value, the test results, as summarized in Table~\ref{table:results}, indicate that GRBF(A1) and parametric-Tanh(A2) deliver lower relative errors.


Fig.~\ref{fig:diffusion} presents the solution of the Convection-Diffusion equation for the velocity and the force values at the beginning and the end of the simulation using parametric Tanh(A2) and the GRBF(A1) models. The reference velocity is concentrated at the center (0.5, 0.5) at the initial time step (initial condition), dissipating as time progresses (as shown by the shrinking central area at time step 1). Similarly, the force field is shown in the third and fourth columns, with the high-intensity areas at time-step zero diffusing over time.

In the second row of Fig.~\ref{fig:diffusion}, the GRBF model captures the dissipation of the velocity field reasonably well, with only minor errors (as seen in the error plot in row (c)). Similarly, for the force field, the GRBF model captures the general diffusion pattern accurately, with relatively low errors (compared to other models). Parametric Tanh(A2) has a slightly higher relative error for velocity than GRBF(A1).

\subsubsection{Cavity Equation} 
The results in Table~\ref{table:results} show that the B-spline(A2) model achieves the lowest errors, followed by Tanh(A2) when solving the Cavity flow problem.
In contrast, the Fourier, Chebyshev, and Jacobi perform less effectively with errors exceeding 30\% across all models.

Fig.~\ref{fig:loss_history_cavity} shows training convergence history for the Cavity problem. 
The B-spline(A2) model demonstrates the most effective convergence, starting from a relatively high initial loss but rapidly decreasing
However, B-spline(A2) exhibits relatively unstable behaviors, characterized by persistent fluctuations throughout the training process.
Ultimately, B-spline(A2) achieves the lowest final training loss value, followed by Tanh(A2) after 60,000 iterations.


Fig.~\ref{fig:cavity_time_0} presents the solution of the cavity problem, comparing the reference solution to predictions made by B-spline(A2) and Tanh(A2) at the initial Cavity flow time step ($t_0$).
The predicted $u_x$ field by the Tanh(A2) is mostly close to zero, but there are slight deviations at the bottom. However, the $v_y$ field shows deviations from the reference. It exhibits sharp gradients along the top boundary. The smooth nature of the Tanh activation seems to struggle to capture the sharp boundary behavior accurately. The pressure field predicted by the Tanh(A2) has significant deviations, particularly near the top boundary. There are some deviations in the B-spline(A2) model, but the overall structure is closer to the reference solution.

Fig. \ref{fig:cavity_time_99} shows the plots of the reference solution, the Tanh, and the B-spline approximations when solving the cavity problem at the steady state. 
The Tanh(A2) model reasonably approximates $u_x$ field, showing a similar pattern to the reference solution with high velocities near the top boundary. However, the velocity distribution near the bottom of the cavity has more deviation from the reference. The $v_y$ field shows noticeable differences compared to the reference. The pressure field predicted by the Tanh(A2) model shows a significant deviation from the reference, especially at the top boundary.

\subsubsection{Computational Complexity}

Table~\ref{table:results} shows the time required for training in seconds per iteration for each model. 
The number of parameters for MLP models is much lower than that of KAN models for basis functions (Table~\ref{table:network_architecture}) and possibly extra weights for handling those functions 
and their derivatives for propagation and physics loss term calculations (Table~\ref{table:model_formulations}). MLP with Tanh activation functions, followed by GRBF, generally have lower times per iteration (t/it), while KAN with Jacobi basis function tends to have the highest computational cost compared to other models. 
\rev{
This increased computational cost for Jacobi can be attributed to its complex polynomial formulation, which requires evaluating higher-degree terms (up to $d=4$) with multiple coefficients that scale with the polynomial degree as shown in Table~\ref{table:model_formulations}.
}

The relatively small increase in time per iteration, despite the larger parameter count in architecture A2 compared to architecture A1 within the same model, is likely a result of the NVIDIA A100's parallel processing capabilities and efficient utilization of optimizations in PyTorch. Furthermore, 128 batch size balances computational efficiency and memory usage. 
Since the same minibatch size is used for both A1 and A2 architectures, the increased parameter count in A2 may not be sufficient to cause significant changes in time per iteration.

\rev{
\section{Spectral Bias and Convergence Analysis}
\label{sec:analysis}
}

\begin{figure*}[t]
\centering
\begin{tabular}{cc}
     \multicolumn{2}{c }{\includegraphics[width=0.47\columnwidth, trim={0cm .0cm 0cm 0cm}, clip]{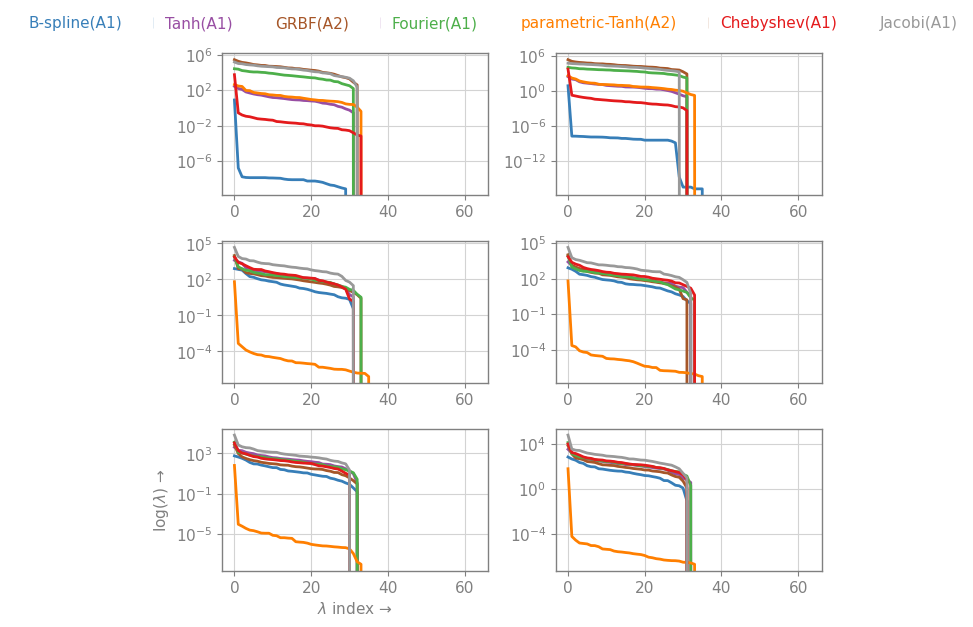}} \\
    \multicolumn{2}{c}{
            \tiny{(a) Helmholtz equation. Left: ${bc}$, right:${phy}$} 
            }
            \\
    \includegraphics[width=0.47\columnwidth, trim={0cm .0cm 0cm 0cm}, clip]{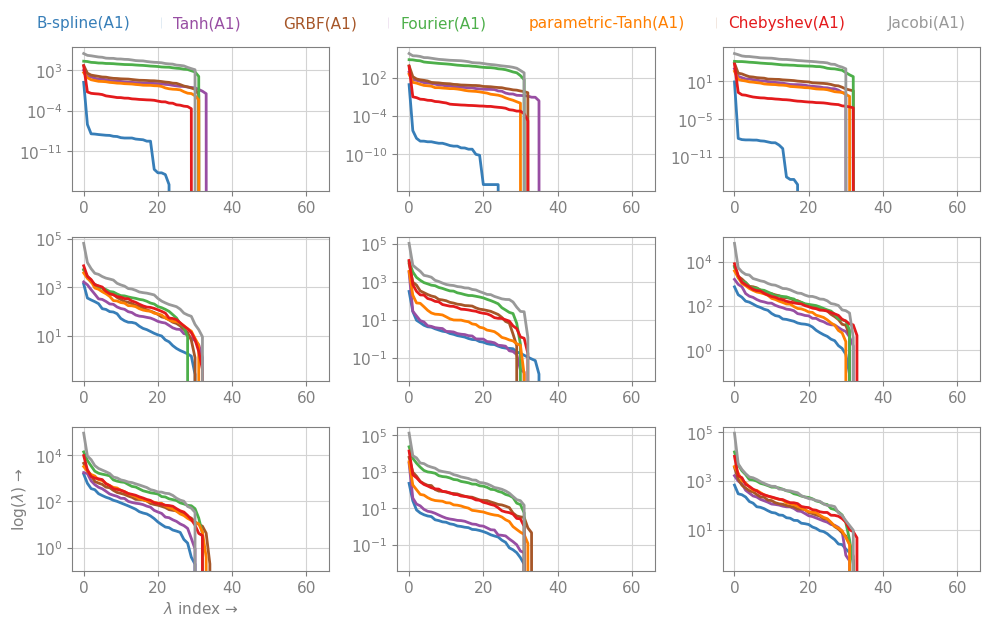}&
    \includegraphics[width=0.47\columnwidth, trim={0cm .0cm 0cm 0cm}, clip]{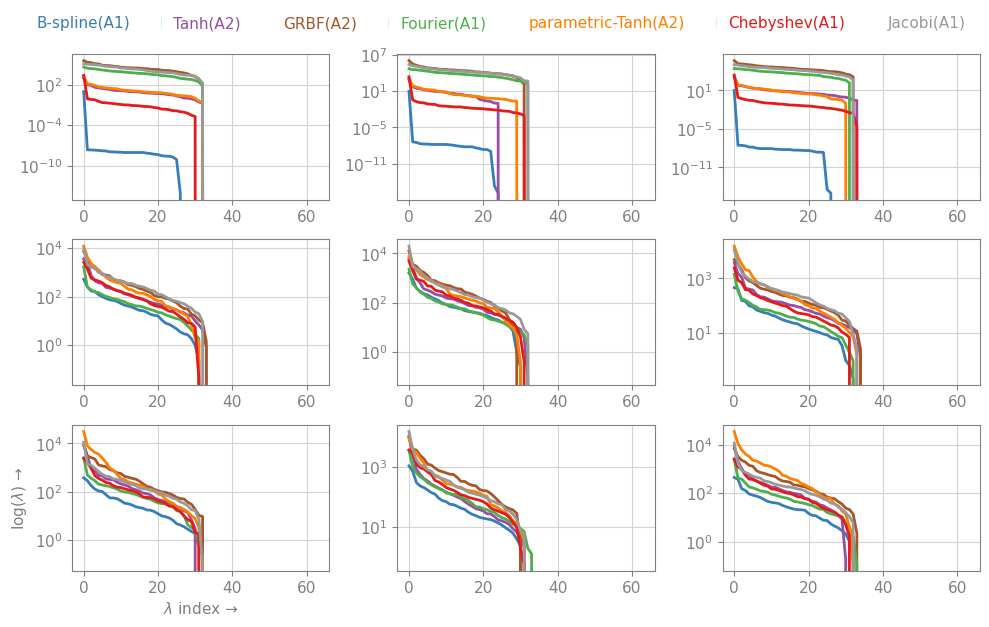} \\

    \tiny{(b) Klein-Gordon equation. Left: ${bc}$, middle:  ${ic}$, right:${phy}$ } &\tiny{(c) Wave equation. Left: ${bc}$, middle:  ${ic}$, right:${phy}$ } \\

    \includegraphics[width=0.47\columnwidth, trim={0cm .0cm 0cm 0cm}, clip]{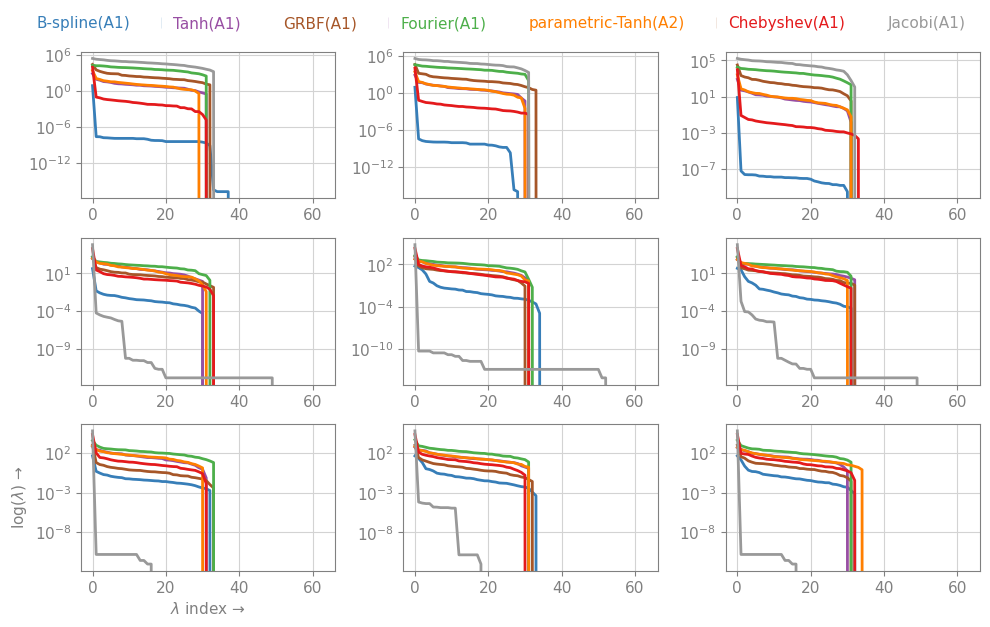} &
   \includegraphics[width=0.47\columnwidth, trim={0cm .0cm 0cm 0cm}, clip]{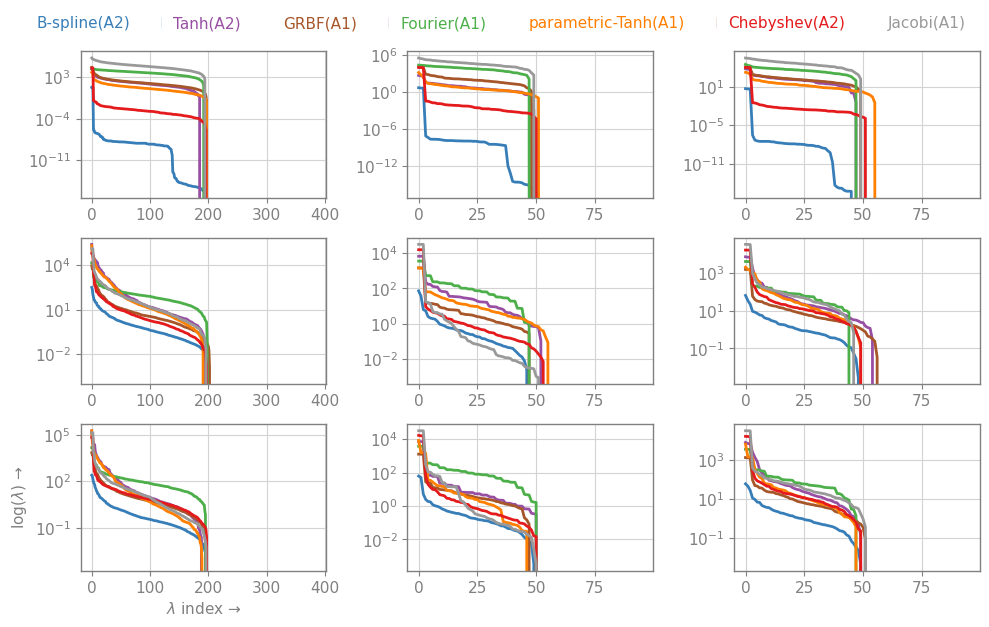}\\
    \tiny{(d) Convection-diffusion equation. Left: ${bc}$, middle:  ${ic}$, right:${phy}$ } &
   \tiny{(e) Cavity equation. Left: ${bc}$, middle:  ${ic}$, right:${phy}$ }\\
    
\end{tabular}
    
\caption{
    \rev{Evolution of the NTK eigenvalue spectra during training, including the zero eigenvalues. 
    Each row of a subplot corresponds to a different training iteration: the initial (top), middle, and final (bottom) stages of the training epochs. The horizontal axis of each plot represents the eigenvalue index, and the vertical axis ($log(\lambda)$) represents the eigenvalue.
    }
}        
\label{fig:ntk_eigenvalues}
\end{figure*}

 \begin{figure*}[t]
    \centering
    \begin{tabular}{c}

        \includegraphics[width=0.65\columnwidth, trim={0cm .30cm 0cm 0cm}, clip]{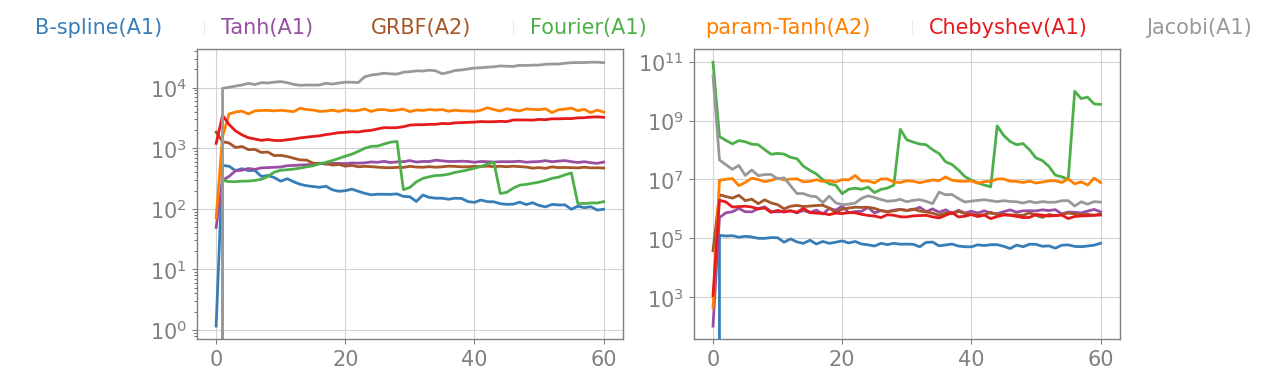} \\

        \tiny{(a) Helmholtz equation. Left: $\lambda_{max}(\nabla^2_{\theta} \mathcal{L}_{bc})$, right: $\lambda_{max}(\nabla^2_{\theta} \mathcal{L}_{phy})$} \\
        
        \includegraphics[width=0.75\columnwidth, trim={0cm .30cm 0cm 0cm}, clip]{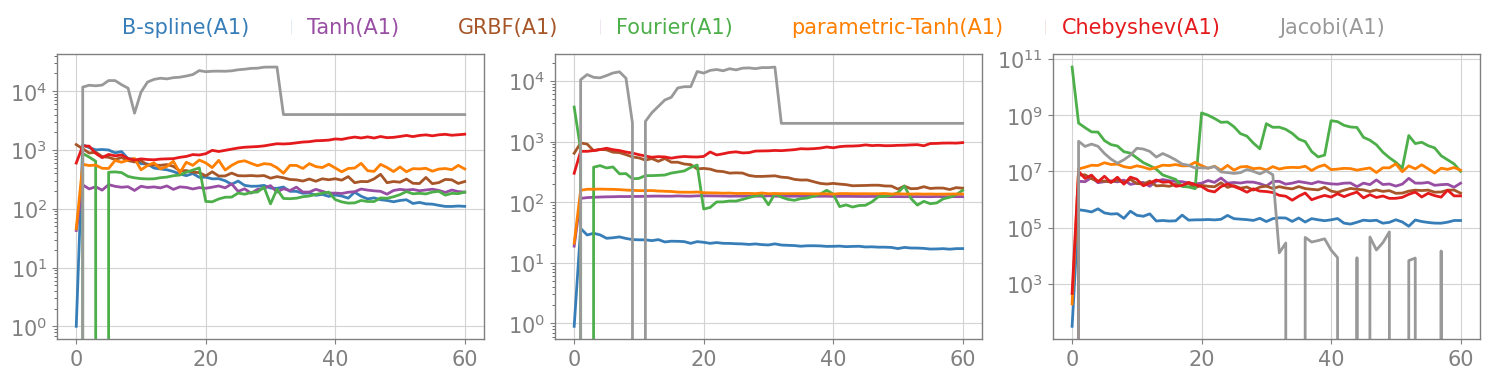} \\
         
        \tiny{(b) Klein-Gordon equation. Left: $\lambda_{max}(\nabla^2_{\theta} \mathcal{L}_{bc})$, middle: $\lambda_{max}(\nabla^2_{\theta} \mathcal{L}_{ic})$, right: $\lambda_{max}(\nabla^2_{\theta} \mathcal{L}_{phy})$} \\
        
        \includegraphics[width=0.75\columnwidth, trim={0cm .30cm 0cm 0cm}, clip]{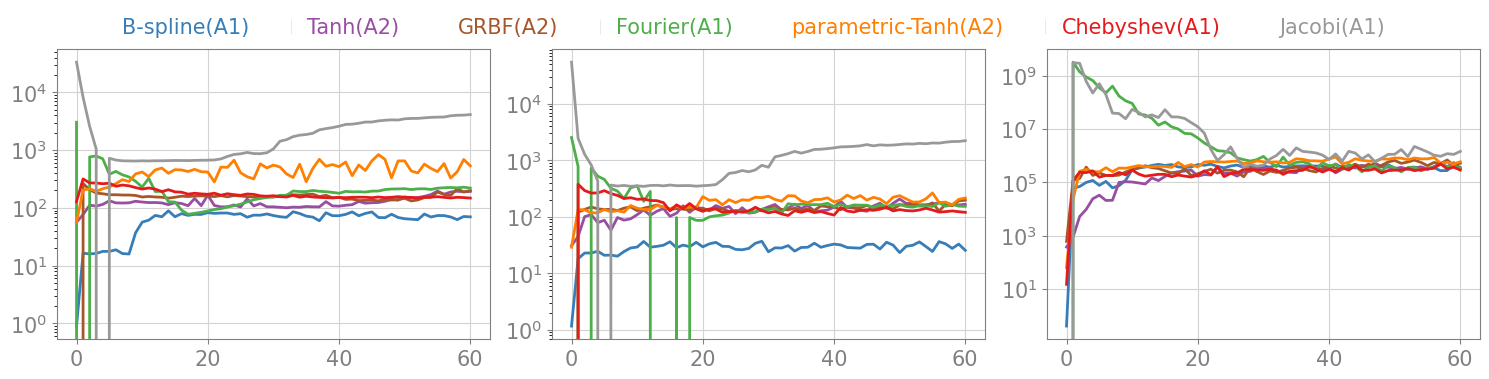} \\
        
        \tiny{(c) Wave equation. Left: $\lambda_{max}(\nabla^2_{\theta} \mathcal{L}_{bc})$, middle: $\lambda_{max}(\nabla^2_{\theta} \mathcal{L}_{ic})$, right: $\lambda_{max}(\nabla^2_{\theta} \mathcal{L}_{phy})$} \\
        
        \includegraphics[width=0.75\columnwidth, trim={0cm .30cm 0cm 0cm}, clip]{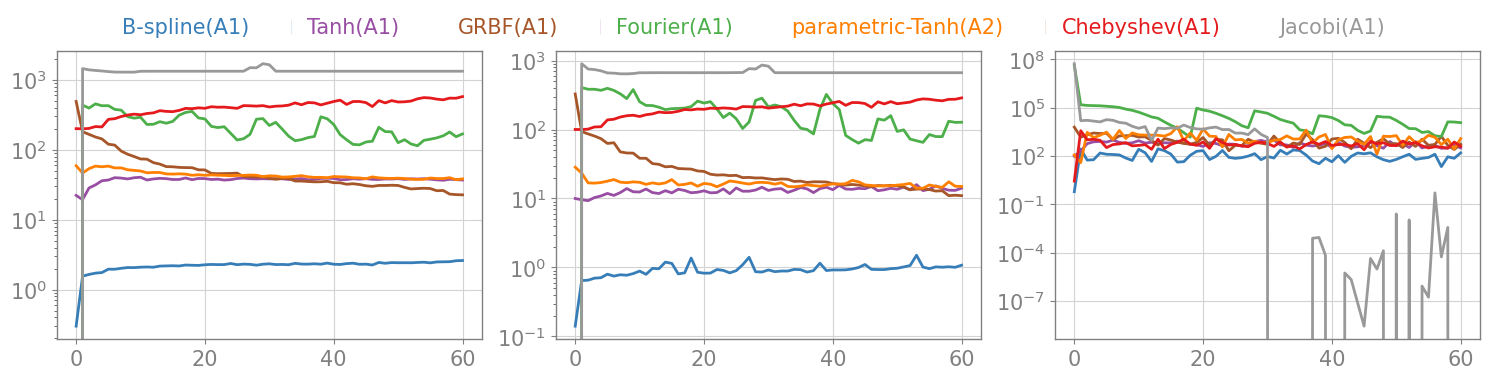} \\
                
        \tiny{(d) Convection-diffusion equation. Left: $\lambda_{max}(\nabla^2_{\theta} \mathcal{L}_{bc})$, middle: $\lambda_{max}(\nabla^2_{\theta} \mathcal{L}_{ic})$, right: $\lambda_{max}(\nabla^2_{\theta} \mathcal{L}_{phy})$} \\
        
        \includegraphics[width=0.75\columnwidth, trim={0cm .30cm 0cm 0cm}, clip]{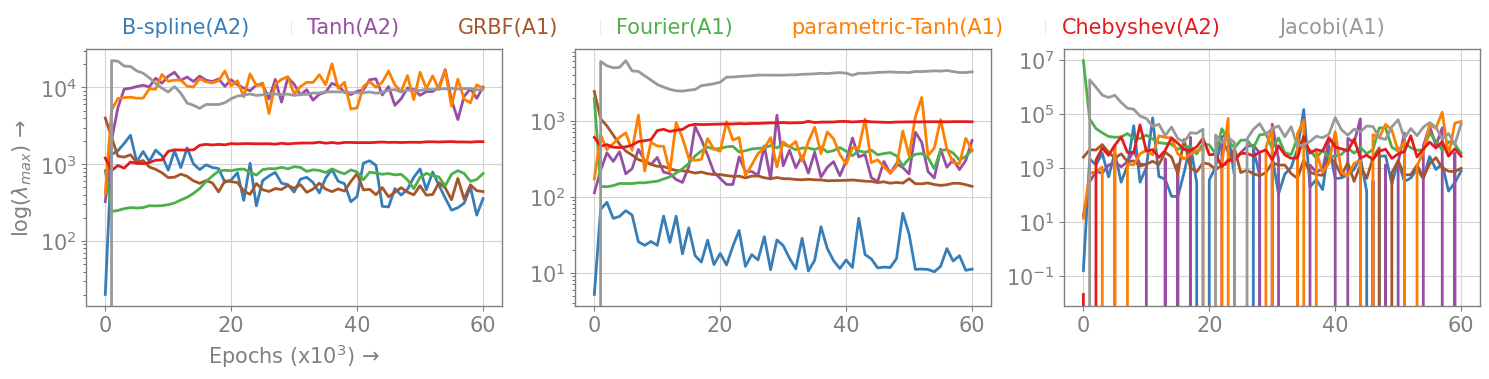} \\

        \tiny{(e) Cavity equation. Left: $\lambda_{max}(\nabla^2_{\theta} \mathcal{L}_{bc})$, middle: $\lambda_{max}(\nabla^2_{\theta} \mathcal{L}_{ic})$, right: $\lambda_{max}(\nabla^2_{\theta} \mathcal{L}_{phy})$} \\
        
    \end{tabular}
    
    \caption{
    Evolution of the maximum eigenvalue of the Hessian of the loss function, $\lambda_{\text{max}}$, over training epochs for different neural network models. Large Hessian eigenvalues may indicate optimization difficulties with high frequencies, especially in the physics loss term.
    }
        
    \label{fig:spectral}
\end{figure*}

In this section, we analyze the behavior of the eigenvalues of \rev{the NTK matrix as a spectral bias measurement and } the maximum Hessian eigenvalue of the loss function with respect to neural network parameters as a convergence measurement tool using the best-performing models from the previous sections. These experiments employ the Adam optimizer, configured with a learning rate of $0.001$ and a weight decay of $10^{-6}$. Additionally, mini-batch gradient descent is employed with a batch size of 128 for a total of 60,000 iterations. 
\rev{
To better understand the behavior of KAN models, only the basic basis function is used; additional activation functions, such as SiLU from Eq.~\ref{eq:bspline_silu}, have been omitted.
}


\rev{
\subsection{Spectral Bias with NTK}
\label{sec:spectral_bias_analysis}
The empirical NTK offers an approximate theoretical framework for understanding the dynamics of neural network training and spectral bias. For a neural network with parameters $\theta$, the empirical NTK matrix between inputs $\mathbf{x}$ and $\mathbf{x}'$ at time $t$ is defined as~\cite{rahaman2019spectral,wang2021understanding}:
\begin{equation}
K_t(\mathbf{x}, \mathbf{x}') = \nabla_\theta f(\mathbf{x}; \theta_t) \cdot \nabla_\theta f(\mathbf{x}'; \theta_t)^\top
\end{equation}
In the context of PINNs, we compute the empirical NTK matrix for batches of inputs $
\mathbf{x}_1 = \{\mathbf{x}_1^{(1)}, \ldots, \mathbf{x}_1^{(n)}\}$ and $\mathbf{x}_2 = \{\mathbf{x}_2^{(1)}, \ldots, \mathbf{x}_2^{(n)}\}$. We construct the Jacobian matrices $\mathbf{J}_1 \in \mathbb{R}^{n \times |\theta|}$ and $\mathbf{J}_2 \in \mathbb{R}^{n \times |\theta|}$ where:
}

\rev{
\begin{equation}
    [\mathbf{J}_1]_{i,j} = \frac{\partial f(\mathbf{x}_1^{(i)}; \theta)}{\partial \theta_j}, \quad [\mathbf{J}_2]_{i,j} = \frac{\partial f(\mathbf{x}_2^{(i)}; \theta)}{\partial \theta_j}
\end{equation}
The empirical NTK matrix is then computed as:
\begin{equation}
    \mathbf{K} = \mathbf{J}_1 \mathbf{J}_2^\top \in \mathbb{R}^{n \times n}
    \label{eq:ntk_formula}
\end{equation}
Following the NTK theory framework~\cite{jacot2018neural,lee2019wide}, the training dynamics in the infinite-width limit can be characterized by analyzing the eigenvalue decomposition of $\mathbf{K}$:
}

\rev{
\begin{equation}
    \mathbf{K} = \mathbf{Q} \boldsymbol{\Lambda} \mathbf{Q}^\top
\end{equation}
\noindent assuming $\mathbf{K}$ is positive semi-definite from Eq.~\ref{eq:ntk_formula}; where, $\boldsymbol{\Lambda} = \text{diag}(\lambda_1, \lambda_2, \ldots, \lambda_n)$ contains the eigenvalues in descending order $\lambda_1 \geq \lambda_2 \geq \cdots \geq \lambda_n \geq 0$ and $\mathbf{Q}$ is orthogonal matrix. 
Since $e^{-\mathbf{K}t}= \mathbf{Q}^\top e^{-\boldsymbol{\Lambda}t} \mathbf{Q}$, the eigenvalue spectrum directly governs the learning dynamics through the exponential convergence rates.
Under gradient descent with learning rate $\eta$, the network function evolves according to:
}

\rev{
\begin{align*}
    \mathcal{L}(\theta) &= \frac{1}{2}\|f(\mathbf{x}; \theta) - f^*(\mathbf{x})\|^2\\
    \frac{d\theta}{dt} &= -\eta \nabla_\theta \mathcal{L}(\theta)\\
    \nabla_\theta \mathcal{L} &= \nabla_\theta f(\mathbf{x}; \theta) \cdot (f(\mathbf{x}; \theta) - f^*(\mathbf{x}))\\
    \frac{df(\mathbf{x}; \theta)}{dt} &= \nabla_\theta f(\mathbf{x}; \theta) \cdot \frac{d\theta}{dt}\\
    \frac{df(\mathbf{x}; \theta)}{dt} &= -\eta \nabla_\theta f(\mathbf{x}; \theta) \cdot \nabla_\theta \mathcal{L}\\
    \frac{df(\mathbf{x}; \theta)}{dt} &= -\eta \nabla_\theta f(\mathbf{x}; \theta) \cdot \nabla_\theta f(\mathbf{x}'; \theta) \cdot (f(\mathbf{x}'; \theta) - f^*(\mathbf{x}'))\\
    \frac{df(\mathbf{x}; \theta)}{dt} &= -\eta \sum_{\mathbf{x}'} K(\mathbf{x}, \mathbf{x}') (f(\mathbf{x}'; \theta) - f^*(\mathbf{x}'))\\
    \frac{df(t)}{dt} &= -\eta \nabla_\theta \mathcal{L}(\theta) \cdot \nabla_\theta f = -\eta \mathbf{K}(f(t) - f^*)
\end{align*}
This is a first-order PDE with the solution:
}

\rev{
\begin{equation}
    f(t) - f^* = e^{-\eta \mathbf{K} t}(f(0) - f^*)
\end{equation}
\noindent where, $f{(t)}$ represents the network predictions at iteration $t$, $f^*$ is the target function, $\eta$ is the learning rate, and $f(0)$ is the initial condition. 
}

\rev{
\begin{equation}
    [\mathbf{Q}^\top(f{(t)} - f^*)]_i \approx e^{-\eta\lambda_i t} [\mathbf{Q}^\top(f{(0)}   - f^*)]_i
    \label{eq:ntk_grad}
\end{equation}
\noindent where, $f{(t)}$ represents the network predictions at iteration $t$, $f^*$ is the target function, and $\eta$ is the learning rate. This reveals that eigenfunction components with larger eigenvalues $\lambda_i$ converge exponentially faster at rate $\eta\lambda_i$.
}

\rev{
For KAN networks with learnable  basis function coefficients $c_{ijr}$, the NTK analysis can be extended:
\begin{equation}
    K^{\text{KAN}}_t(\mathbf{x}, \mathbf{x}') = \sum_{i,j,r} \frac{\partial f(\mathbf{x})}{\partial c_{ijr}} \frac{\partial f(\mathbf{x}')}{\partial c_{ijr}} \bigg|_{\theta_t}
\label{eq:kan_ntk}
\end{equation}
Thus, the choice of the basis function significantly influences the NTK eigenvalue spectrum and resulting spectral bias. However, the user-defined variables, such as grid size and basis function order, can make the analysis less straightforward. 
}

\rev{
In general, it is proven that Fourier-based activations modify the NTK to achieve an improved high-frequency representation, resulting in eigenvalue spectra with slower decay~\cite{tancik2020fourier,xu2025understanding}. For GRBF kernels, higher eigenvalues correspond to low-frequency (smooth) functions, while lower eigenvalues correspond to high-frequency (oscillatory) functions~\cite{jacot2018neural,rahaman2019spectral}. B-spline and polynomial activations exhibit intermediate behavior, with eigenvalue distributions that depend on the order of the basis function and the domain coverage\cite{jagtap2020adaptive,liang2021reproducing,xu2025understanding}.
}

\rev{
Fig.~\ref{fig:ntk_eigenvalues} shows the evolution of the NTK eigenvalue spectra in descending order for the initial (top), middle (middle), and end (bottom)
of training iteration. The figure reveals that even with adaptive basis functions, spectral bias persists to different degrees. 
}
\rev{
The NTK spectra show that the training dynamics evolve in a problem-dependent manner. At initialization (top row), Fourier and Jacobi activations consistently exhibit the flattest eigenvalue decay across all problems, indicating their capacity to represent high-frequency modes. In contrast, B-spline and Tanh activations show steep initial decay. As training progresses (middle and bottom rows), all models experience varying degrees of spectral compression, reflecting a gradual collapse of higher NTK eigenvalues. However, Fourier maintains a broader spectrum throughout training. While the NTK spectrum suggests that B-splines exhibit high spectral bias, they perform remarkably well on testing data. This can be understood by drawing an analogy to finite basis PINNs (FBPINNs)~\cite{moseley2023finite,dolean2024multilevel}, where domain decomposition ensures that globally high-frequency content appears locally low-frequency. Similarly, B-splines, being piecewise-defined with local support,  can capture local oscillations, circumventing global spectral limitations.
}

\rev{
Examining the NTK eigenvalues (plots from left to right in Fig.~\ref{fig:ntk_eigenvalues}) reveals essential insights into the learning dynamics across loss components. For the boundary condition domain, most activations show rapid eigenvalue decay due to the inherent smoothness of boundary enforcement. The initial condition domain presents an intermediate decay profile, balancing smoothness with the need to capture initial transients. The physics residual domain consistently exhibits the broadest NTK spectra, primarily due to its larger dataset size compared to the boundary and initial components. Notably, activations such as Fourier and Jacobi consistently retain higher eigenvalues in the physics domain, ensuring their capacity to capture high-frequency features and complex solution structures that arise within the domain. 
}

\rev{
In terms of training dynamics, the NTK eigenvalue spectra initially reflect the inherent biases of each activation architecture. As training progresses, most activations, except Fourier, demonstrate progressive spectral compression, indicating a collapse towards lower frequencies. Notably, Fourier activations maintain a broader spectrum throughout training. 
}

\subsection{Convergence with Loss Hessian}

\rev{
We extend the NTK analysis by examining the maximum eigenvalue of the Hessian matrix $\mathbf{H}$ to analyze the convergence behavior of the models~\cite{ghorbani2019investigation, cohen2021gradient, alain2019negative, liao2021hessian, foret2020sharpness,kaur2023maximum}. We utilized the best-performing models discussed in Section~\ref{sec:results}. To minimize the computation overhead, we estimate the largest eigenvalue $ \lambda_{\max}$ of the Hessian matrix as a Rayleigh quotient using the power iteration method~\cite{montavon2012neural,yao2020pyhessian}:
}

\rev{
\begin{equation}
    \lambda_{\max} = \lim_{k \rightarrow \infty} \frac{\mathbf{v}^T\mathbf{H}^k\mathbf{v}}{\mathbf{v}^T\mathbf{H}^{k-1}\mathbf{v}}
\end{equation} 
\noindent where, $k$ is the iteration number, $\mathbf{H}$ is the Hessian matrix, and $\mathbf{v}$ is a random initial vector that converges to the eigenvector corresponding to the maximum eigenvalue.
}

Fig.~\ref{fig:spectral} presents the maximum eigenvalue of the Hessian of the loss function, denoted as $\lambda_{max}(\nabla^2_{\theta} \mathcal{L})$.
The B-spline basis function consistently exhibits lower and more stable values of $\lambda_{max}$, which indicates a smoother optimization process. This observation is supported by the results discussed earlier, where B-splines achieve low error rates and demonstrate steady convergence across various test cases.

The GRBF basis functions initially exhibit some instability as indicated by the fluctuating $\lambda_{max}$, but eventually stabilize. While the Fourier kernel functions are naturally well-suited for representing periodic, high-frequency components, the observed fluctuations in $\lambda_{max}$ indicate some optimization challenges due to sharp curvatures in the loss landscape.  Similarly,  Chebyshev and Jacobi basis functions show significant instability, as reflected in their fluctuating $\lambda_{max}$ values. This instability is confirmed by their higher error rates and unstable convergence seen in the loss history figures. In the case of the Jacobi basis function, the significant fluctuations in $\lambda_{max}$, particularly within the physics loss term, indicate major instability, which correlates with poor performance in several test cases.

Fig.~\ref{fig:spectral} shows that the maximum eigenvalue of the physics loss consistently takes longer to stabilize and is generally larger than the eigenvalues of the boundary and initial condition losses, even for more stable basis functions like B-splines. 
This discrepancy occurs because the physics loss Hessian captures more complex, high-frequency information and is applied across the entire problem domain, often utilizing more training datasets than the other terms. Additionally, the physics loss term has lower weighting values compared to the higher weighting values of the boundary and initial loss terms, as discussed in Section~\ref{sec:use-cases}.

In cases, e.g., Helmholtz, Klein-Gordon, and Cavity problems, the fluctuations in the physics loss Hessian are particularly pronounced. The presence of negative eigenvalues in the physics loss Hessian, as observed in Convection-diffusion and Cavity problems, highlights the challenging optimization landscape for the physics loss compared to the boundary and initial condition losses. The appearance of negative curvature further indicates that gradient-based optimizers, such as Adam, may struggle in these cases, as they are better suited for convex optimization problems. Negative curvature can cause optimization algorithms to get trapped in saddle points or lead to slower convergence.
Recent studies, such as~\cite{rathore2024challenges}, suggest Newton's method or preconditioned gradient descent techniques can help navigate these regions by using second-order information to escape saddle points and improve convergence rates.

\section{Conclusion}

\rev{
Our results show a trade-off between expressivity and training convergence stability in learnable activation functions. While these functions demonstrate notable effectiveness in simpler architecture, A1, they face significant challenges when scaled to complex networks, A2. This limitation stems from the curse of functional dimensionality inherent to KAN architectures: unlike MLPs that learn feature hierarchies through weight matrices, KANs must learn entire function spaces at each connection. This creates an exponentially expanding optimization landscape that overwhelms gradient-based methods. The architectural complexity ultimately leads to optimization intractability and performance degradation across multiple PDE types. This explains the instability of learnable activations, including the parametric-Tanh function, which introduces only two learnable parameters and experiences similar scaling challenges, albeit to a lesser degree.
}

\rev{
Contrary to common intuition, low spectral bias does not necessarily guarantee better testing accuracy. A notable example of this is the Fourier basis function, which has been proven to improve spectral bias properties.
Our analysis shows that while Fourier and Jacobi basis functions maintain broader NTK eigenvalue spectra throughout the training, they simultaneously exhibit the largest maximum Hessian eigenvalues in physics residual and boundary components, suggesting a potential convergence instability. Conversely, B-splines display rapid NTK eigenvalue decay, indicating high spectral bias. However, their distinct advantage lies in maintaining high-frequency representation within localized support regions, effectively capturing sharp gradients without introducing global high-frequency interference.
}

\rev{
Our findings emphasize the significance of problem-specific activation selection. The Tanh activation function is particularly effective for wave-like PDEs, such as the Klein-Gordon and wave equations. Surprisingly, the parametric-Tanh fails on the Wave equation. This underscores that activation function selection remains inherently problem-dependent. B-splines prove particularly effective for PDEs with localized features, leveraging their piecewise polynomial construction to model sharp gradients and boundary layers, as demonstrated in the Cavity flow problem. In contrast, Fourier bases struggle with sharp transitions, producing oscillatory artifacts near discontinuities, highlighting the need for advanced optimization techniques or better hyperparameter tuning.
}

Beyond activation choices, the relative weighting of physics, boundary, and initial condition loss terms remains critical for training success across all architectures. Our Hessian eigenvalue analysis reveals consistently large and unstable eigenvalues, particularly in physics loss terms, indicating that modifications to activation and basis functions alone cannot resolve the fundamental optimization challenges of PINN loss landscapes. Future research should address these optimization issues while developing theoretical frameworks to predict optimal spectral bias distributions for different PDE classes. In conclusion, this work enhances our understanding of how different architectural choices and activation functions impact the performance of neural PDE solvers.

\section*{Acknowledgment}
We thank the National Center for High-Performance Computing of Turkey (UHeM) for providing computing resources under grant number 5010662021. 
We thank Dr. Saiful Khan from Rutherford Appleton Laboratory, Science and Technology Facilities Council, UK, for his valuable comments and extensive feedback on the manuscript. We also thank Emre Cenk Ersan for generating the Cavity flow problem using Ansys Fluent software. 


\bibliographystyle{unsrt}  
\bibliography{references}  

\end{document}